\documentclass[twoside]{article}

\usepackage{aistats2024_fsp}

\usepackage[round]{natbib}

\usepackage{graphicx}
\usepackage{url}
\usepackage{booktabs}
\usepackage{amsfonts}
\usepackage{amsmath}
\usepackage{amsthm}
\usepackage{enumerate}
\usepackage{mathtools}
\mathtoolsset{showonlyrefs}
\usepackage{bm}
\PassOptionsToPackage{hyphens}{url}
\usepackage[hyperfootnotes=false]{hyperref}

\theoremstyle{definition}
\newtheorem{definition}{Definition}[section]
\newtheorem{proposition}[definition]{Proposition}

\DeclareMathOperator*{\argmin}{argmin}
\DeclareMathOperator*{\argmax}{argmax}
\newcommand{\bfx}{\mathbf{x}}
\newcommand{\bfX}{\mathbf{X}}
\newcommand{\bfz}{\mathbf{z}}
\newcommand{\bfZ}{\mathbf{Z}}
\newcommand{\bfy}{\mathbf{y}}
\newcommand{\bfY}{\mathbf{Y}}
\newcommand{\R}{\mathbb{R}}

\runningauthor{September, Sanna Passino, Goldmann and Hinel}
\runningtitle{Extended Deep Adaptive Input Normalization for Preprocessing Time Series Data}
\usepackage{fontawesome,comment}
\usepackage[hang,flushmargin]{footmisc}
\begin{document}

\runningtitle{Extended Deep Adaptive Input Normalization for Preprocessing Time Series Data}

\runningauthor{September, Sanna Passino, Goldmann and Hinel}

\twocolumn[

\aistatstitle{
    % Title idea 1
    Extended Deep Adaptive Input Normalization for
    Preprocessing Time Series Data for
    Neural Networks
}

\aistatsauthor{
    Marcus A. K. September
    \And Francesco Sanna Passino
}

\aistatsaddress{
    Department of Mathematics \\ Imperial College London
    \And Department of Mathematics \\ Imperial College London
}%

\aistatsauthor{
    Leonie Goldmann%
    \And Anton Hinel%
}

\aistatsaddress{
    Decision Science, Credit \& Fraud Risk \\ American Express
    \And Decision Science, Credit \& Fraud Risk \\ American Express
}%
]

\begin{abstract}
  % Introduce data preprocessing 
  Data preprocessing is a crucial part of any machine learning pipeline,
  and it can have a significant impact on both performance and training efficiency.
  This is especially evident when using deep neural networks for time series prediction and
  classification:
  real-world time series data often exhibit irregularities such
  as multi-modality, skewness and outliers, and the model performance can degrade 
  rapidly if these characteristics are not adequately addressed.
  % Pivot towards the work of this paper
  In this work, we propose the EDAIN (Extended Deep Adaptive Input Normalization) layer, a novel adaptive neural layer that 
  learns how to appropriately normalize irregular time series data
  % Summarise how EDAIN works
  for a given task in an end-to-end fashion, instead of using a fixed normalization scheme.
  This is achieved by optimizing its unknown parameters simultaneously with the deep neural network
  using back-propagation.
  % Summarise the proposeed method's effectiveness
  Our experiments, conducted using synthetic data, a credit default prediction dataset, and a large-scale limit order book benchmark dataset,
  % A sentence about the results
  demonstrate the superior performance of the EDAIN layer when compared to conventional normalization methods and existing adaptive time series
  preprocessing layers.
\end{abstract}

  \blfootnote{\noindent Corresponding author: Francesco Sanna Passino \\ \faEnvelopeO\ \texttt{f.sannapassino@imperial.ac.uk} \\[0.5em]
  %This work was completed while Marcus A. K. September was a master's student at Imperial College London. \\[0.5em]
  For the purpose of open access, the authors have applied a Creative Commons Attribution (CC-BY) licence to any \textit{Author Accepted Manuscript} version arising.
}

%%%%%%%%%%%%%%%%%%%%%%%%%%%%%%%%%%%%%%%%%%%%%%%%%%%%%%%%%%%%
\section{INTRODUCTION}%
\label{sec:intro}

There are many steps required when applying deep neural networks or, more generally, any machine learning model, to a problem. First, data should be gathered, cleaned, and formatted into machine-readable values. Then, these values need to be preprocessed to facilitate learning. Next, features are designed from the processed data, and the model architecture and its hyperparameters are chosen. This is followed by parameter optimisation and evaluation using suitable metrics. These steps may be iterated several times.

A step that is often overlooked in the literature is \textit{preprocessing} \citep[see, for example,][]{stanislav}, which consists in operations used to transform raw data in a format that is suitable for further modeling, such as detecting outliers, handling missing data and normalizing features. Applying appropriate preprocessing to the data can have significant impact on both performance and training efficiency \citep{brits,nawi, dain,preprocess_origin,bin}. However, determining the most suitable preprocessing method usually requires a substantial amount of time and relies on iterative training and performance testing. Therefore, the main objective of this work is to propose a \textit{novel efficient automated data preprocessing method for optimising the predictive performance of neural networks}, with a focus on normalization of multivariate time series data.

\subsection{Preprocessing multivariate time series}

% DONE: introduce notation
Let $\mathcal{D}=\left\{\bfX^{(i)} \in  \R^{d \times T},\ i=1,\dots,N\right\}$ denote a dataset
containing $N$ time series, where each time series $\bfX^{(i)} \in \R^{d \times T}$ is composed of
$T$ $d$-dimensional feature vectors. The integers $d$ and $T$ refer to the
feature and temporal dimensions of the data, respectively.
Also, we use $\bfx^{(i)}_t \in \R^d,\ t=1,\dots,T$ to
refer to the $d$ features observed at timestep $t$ in the $i$-th time series.
% DONE: general motivation for more sophisticated preprocessing methods
%Consider a dataset of time series on the form $\mathcal{D}=\left\{ \bfX^{(i)} \in \R^{d \times T},\ i=1,\dots,N\right\}$, where $\bfx^{(i)}_t \in \R^d$ denotes the vector for features
%recorded at timestep $t$ for time series $i$.
%In time series analysis, i
Before feeding the data into a model such as a deep neural network, it is common for practitioners to perform $z$-score normalization \citep[see, for example,][]{stanislav} on
$\bfx^{(i)}_t = (x^{(i)}_{t,1},\dots,x^{(i)}_{t,d}) \in \R^d$,
obtaining
\begin{equation}
   \tilde{x}^{(i)}_{t,k} = \frac{x^{(i)}_{t,k} - \mu_k}{\sigma_k},\ k=1,\dots,d,
\end{equation}
where $\mu_k$ and $\sigma_k$ denote the mean and standard deviation of the measurements
from the $k$-th predictor variable. 
Another commonly used method is min-max scaling, where the
observations for each predictor variable are transformed to fall in the value range $[0,1]$ \citep[see, for example,][]{stanislav}. 
%This type of normalization is particularly appropriate when the observations follow a Gaussian distribution. 
Other common preprocessing methods are
winsorization \citep[see, for example,][]{winsorization} and the Yeo-Johnson power
transform \citep[see, for example,][]{yeoJohnson}.
In this work, we refer to these conventional methods as \textit{static preprocessing} methods, as the transformation parameters are fixed 
statistics that are computed through a single sweep of the training 
data. 
Most of these transformations only change the location and scale of the observations, but 
%However, this assumption holds extremely rarely in 
%practice, as 
real-world
data often contains additional irregularities such as skewed distributions, outliers, extreme values,
heavy tails and multiple modes \citep{mixture_ct, nawi}, which are not mitigated by transformations such as $z$-score and min-max normalization. %, potentially leading to sub-optimal results. %and requiring more sophisticated preprocessing methods.
Employing static normalization in such cases may lead to sub-optimal results, as demonstrated in \cite{dain,rdain,bin} and in the experiments on real and synthetic data in \autoref{sec:experimental_evaluation} of this work.

In contrast, better results are usually obtained by employing \textit{adaptive} preprocessing methods \citep[see, for example][]{batchnorm,dain,rdain,bin}, where the preprocessing is integrated into the deep neural network by augmenting its architecture with additional layers. Both the transformation parameters and the
neural network model parameters are then jointly optimised in an end-to-end fashion as part of the objective function of interest.
The main contribution of this this paper belongs to this class of methods: we propose a novel adaptive normalization approach, called EDAIN (Extended Deep Adaptive
Input Normalization), which can appropriately handle irregularities in the input data, without making any assumption on their distribution.

% TODO: define and motiviate the problem/research question.

% TODO: add diagram for local-aware vs. global-aware, and explain the difference

%TODO: Somewhere in introduce, explain what I mean by adaptive: "... these normalization
%approaches are \textit{adaptive} in the sense that their transformation parameters are 
%trained to adapt to the optimisation objective of interest." 

\subsection{Contributions}

% TODO: Briefly explain EDAIN, and emphasize that its the main contribution of this paper

\begin{figure*}[t]
\vspace{.0in}
\centerline{
    % TODO: change figure to rename "local-aware" to "local-aware"
    % TODO: change fonts to cmr (computer modern) and re-export
    \includegraphics[width=\textwidth]{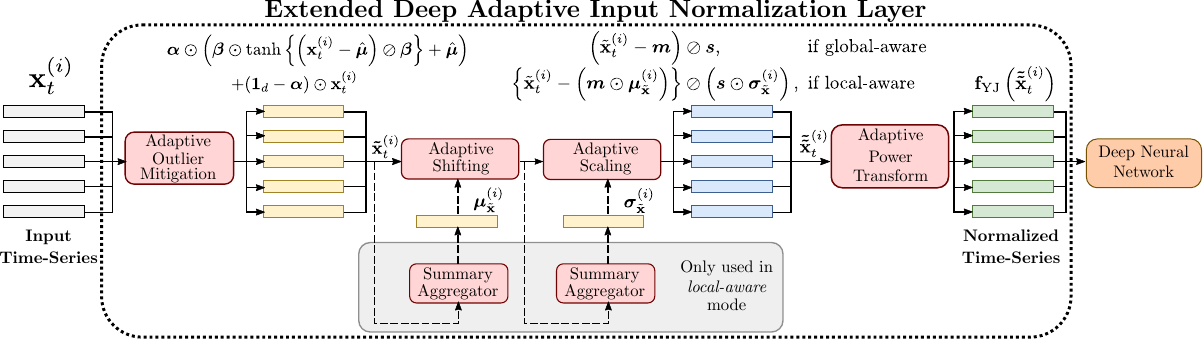}
}
\vspace{.0in}
\caption{
    Architecture of the proposed EDAIN (Extended Deep Adaptive Input Normalization) layer.
    The layout and color choices of the diagram are based on Figure~1 from \cite{dain}.
}%
\label{fig:edain-arch}
\end{figure*}

The main contribution of our work is EDAIN, displayed in \autoref{fig:edain-arch}, a neural layer that can be added to any neural network architecture for preprocessing multivariate time series. This method complements the shift and scale layers proposed in DAIN \citep[described in details in \autoref{sec:related_work}]{dain} with an adaptive outlier mitigation sublayer and a power transform sublayer, used to handle common irregularities observed in real-world data, such as outliers, heavy tails, extreme values and skewness.

Additionally, our EDAIN method can be implemented in two versions, named \textit{global-aware} and \textit{local-aware}, suited to unimodal and multi-modal data respectively. 
Furthermore, we propose a computationally efficient variation of EDAIN, trained via the Kullback-Leibler divergence, named EDAIN-KL. Like EDAIN, this method can normalize skewed data with outliers, but in an unsupervised fashion, and can be used in conjunction with non-neural-network models. 

EDAIN is described in details in \autoref{sec:edain}, after a discussion on related methods in \autoref{sec:related_work}. The proposed methodology is extensively evaluated on synthetic and real-world data in \autoref{sec:experimental_evaluation}, followed by a discussion on its performance and a conclusion. Also, an open-source implementation of the EDAIN layer,
along with code for reproducing the experiments, is available
in the GitHub repository \href{https://github.com/marcusGH/edain_paper}{\texttt{marcusGH/edain\_paper}}.

% Main EDAIN architecture diagram (we want this to appear on page 2)

\section{RELATED WORK}%
\label{sec:related_work}

% DONE: describe more sophisticated methods applied inside NN, like BatchNorm etc.
Several works consider adaptive normalization methods, but they all apply the normalization
transformation to the outputs of the inner layers within the neural network, known as activations in the literature. A well-known example of these transformations is batch normalization \citep{ioffe15}, which applies $z$-score normalization to the output of each inner layer, but several alternatives and extensions exist
\citep[see, for example,][]{huang2020,batchnorm,yu2022}.
% DONE: Introduce DAIN, RDAIN and BIN
To the best of our knowledge, there are only three other methods 
where the deep neural network is augmented by inserting the
adaptive preprocessing layer as the first step, transforming
the data before it enters the network:
the Deep Adaptive Input Normalization (DAIN) layer \citep{dain},
the Robust Deep Adaptive Input Normalization (RDAIN) layer \citep{rdain}, and
the  Bilinear Input Normalization (BIN) layer \citep{bin}.
% We will describe the DAIN layer in more detail in the next paragraph, as this method resembles our proposed EDAIN method the most.

% DONE: detailed description of the DAIN layer
The DAIN layer normalizes each time series $\bfX^{(i)}$ using three sublayers: each time series
is first shifted, then scaled, and finally passed through a
gating layer that can suppress irrelevant features.
The unknown parameters are the weight matrices
$\mathbf{W}_a$, $\mathbf{W}_b$, $\mathbf{W}_c \in \R^{d \times d}$, and the bias
term $\mathbf{d} \in \R^d$, and are used
for the shift, scale, and gating sublayer, respectively. The %adaptive shift layer andadaptive scale layer, 
first two layers together, perform the operation
\begin{equation}\label{eq:dain_tild}
    \tilde{\bfx}_t^{(i)}=\left(
        \bfx_t^{(i)}-\mathbf{W}_a \mathbf{a}^{(i)}
    \right) \oslash \mathbf{W}_b \mathbf{b}^{(i)},
\end{equation}
where $\bfx_t^{(i)} \in \R^{d}$ is the input feature vector at timestep $t$ of time series $i$, $\oslash$ denotes element-wise division, and $\mathbf{a}^{(i)} \in \R^{d}$ and $\mathbf{b}^{(i)} \in \R^d$ are summary statistics that are computed for
the $i$-th time series as follows:
\begin{equation}
    \mathbf{a}^{(i)} =\frac{1}{T} \sum^{T}_{t=1} \bfx_t^{(i)},\ 
    \mathbf{b}^{(i)} = %b_k^{(i)}&=
    \sqrt{
    \frac{1}{T} \sum^{T}_{t=1} \left(\bfx^{(i)}_{t} - \mathbf{W}_a  \mathbf{a}^{(i)} \right)^2}\label{eq:dain_b}.
\end{equation}
In Equation \eqref{eq:dain_b}, the power operations are applied element-wise.
The third sublayer, the gating layer, performs the operation
\begin{equation}
    \tilde{\tilde{\bfx}}_t^{(i)}=\tilde{\bfx}^{(i)}_t \odot S\left( \mathbf{W}_c \mathbf{c}^{(i)} + \mathbf{d}\right).
\end{equation}
Here, $\odot$ is the element-wise multiplication operator,
$S:\mathbb R^d\to\mathbb R^d$ denotes the logistic %sigmoid 
function applied element-wise, and $\mathbf{c}^{(i)}$ is the %third 
summary statistic%, computed as
\begin{equation}
    \mathbf{c}^{(i)}=\frac{1}{T} \sum^{T}_{t=1} \tilde{\bfx}_t^{(i)}.
\end{equation}
The final output of the DAIN layer is thus
\begin{equation}
    \tilde{\tilde{\bfX}}^{(i)}=\left[\tilde{\tilde{\bfx}}^{(i)}_1, \dots, 
\tilde{\tilde{\bfx}}^{(i)}_T\right]\in \R^{d \times T}.
\end{equation}

% Short follow-up mentioning:
% * RDAIN extends DAIN with a residual connection across shift and scale layers
% * BIN performs a similar shift and scale normalization, but with stastistics across both feature-axis and time-axis
% * Both of these methods only consider shift and scale, so not
%   designed to handle other irregularities like heavy tailed distributions, outliers and extreme values.
% * OUR METHOD ....
% DONE: introducing local-aware normalization

% Explaining RDAIN
In the RDAIN layer proposed by \cite{rdain}, a similar 3-stage
normalization pipeline as that of the DAIN layer is used, but
a residual connection across the shift and
scale sublayers is also introduced.
% Explaining BIN
The BIN layer \citep{bin} has two sets of linear shift and scale sublayers
that work similarly to the DAIN layer,
which are applied across columns and rows of each time series $\bfX^{(i)} \in \R^{d \times T}$.
The output of the BIN layer is a trainable linear combination
of the two.

% Explaining KDIT
In addition to the described adaptive preprocessing methods,
using a combination of static preprocessing methods has been proposed.
In particular, \citeauthor{kdit} proposes the
Kernel Density Integral Transformation (KDIT). In KDIT, the data distribution
is estimated via a Gaussian kernel density estimate with bandwidth depending on a parameter $\alpha\in\mathbb R_+$. The estimated density is then used to construct an estimate of the cumulative distribution function, which is used in turn to standardise the data to the range $[0,1]$ \citep{kdit}. 
For $\alpha\to\infty$, KDIT converges to min-max scaling, whereas $\alpha\to0$ corresponds to a quantile transformation \citep{kdit}.

% Briefly reference local-aware property of the three layers, and its
% drawbracks. Then bring up the proposed layer "As we will see in the 
% next section, our proposed ...."
In real world applications, data often present additional irregularities, such as outliers, extreme values, heavy tails and skewness, which the aforementioned adaptive preprocessing methods are not
designed to handle. Therefore, this work proposes EDAIN, a layer which comprises two novel sublayers that can appropriately treat
skewed and heavy-tailed data with outliers and extreme values, resulting in significant improvements in performance metrics on real and simulated data. 
Also, the DAIN, RDAIN and BIN adaptive preprocessing methods are only primarily
designed to handle multi-modal and non-stationary time series,
which are common in financial forecasting tasks \citep{dain}.
% They do this by making the transformation
% parameters a parameterised function of each $\bfX^{(i)}$, allowing a \textit{local-aware}
% transformation that gives a common representation space regardless of the mode of the
% data each measurement originated from. 
They do this by making the shift and scale parameters a parameterised
function of each $\bfX^{(i)}$, allowing a transformation specific to each time series data point, henceforth referred to as \textit{local-aware} preprocessing.
% Introduce EDAIN and its advantages over BIN, DAIN, RDAIN
However, these normalization schemes do not necessarily preserve
the relative ordering between time series data points, which can
degrade performance on unimodal datasets.
As discussed in the next section, we address this drawback by proposing a novel \textit{global-aware} version for our proposed EDAIN layer,
%also supports,
which %eliminates this drawback
preserves ordering
by learning a monotonic transformation. 
It must be remarked that the EDAIN layer can also be fitted in local-aware fashion, to address multi-modality when present, providing additional flexibility when modelling real-world data.

\section{EXTENDED DEEP ADAPTIVE INPUT NORMALIZATION} \label{sec:edain}

%The main contribution of this work 
In this section, we describe in detail the novel EDAIN preprocessing layer, which can be added to any deep learning architecture for time series data. 
EDAIN adaptively applies local transformations specific to each time series, or global transformations across all the observed time series $\bfX^{(i)},\ i=1,\dots,N$ in the dataset $\mathcal{D}$. 
These transformations are aimed at appropriately preprocessing the data, mitigating the effect of skewness, outliers, extreme values and heavy-tailed distributions. 
This section discusses in details the different sublayers of EDAIN, its global-aware and local-aware versions, and training strategies via stochastic gradient descent or Kullback-Leibler divergence minimization. 

% DONE: Reference architecture diagram, and explain how EDAIN works at a high-level
An overview of the EDAIN layer's architecture is shown in \autoref{fig:edain-arch}.
Given some input time series $\bfX^{(i)} \in \R^{d \times T}$, each feature vector
$\bfx^{(i)}_t \in \R^d$ is independently transformed sequentially in four stages:
an outlier mitigation operation $\mathbf{h}_1 : \R^d \rightarrow \R^d$,
a shift operation: $\mathbf{h}_2 : \R^d \rightarrow \R^d$,
a scale operation: $\mathbf{h}_3 : \R^d \rightarrow \R^d$,
and a power transformation operation: $\mathbf{h}_4 : \R^d \rightarrow \R^d$.
% The shift and scale operations also differ based on whether the local-aware or
% global-aware version of the EDAIN layer is used.

\paragraph{Outlier mitigation sublayer.}

In the literature, it has been shown that an appropriate treatment of outliers and extreme values can increase predictive performance
\citep{outlier_wind}. The two most common ways of doing this are omission and winsorization
\citep{winsorization}. The former corresponds to removing the outliers from further analysis, whereas the latter seeks to replace outliers with a censored value corresponding to a given percentile of the observations. %As we want to use all of our data, we only consider the latter.
%Moreover, 
In this work we 
propose the following smoothed winsorization operation obtained via the $\tanh(\cdot)$ function:
\begin{equation}\label{eq:adaptive_windsorization}
\breve{\bfx}_t^{(i)} = \bm\beta \odot
        \tanh\left\{\left(\bfx_t^{(i)}-\hat{\bm\mu}\right) \oslash \bm\beta  \right\}+\hat{\bm\mu},
\end{equation}
where the parameter $\bm \beta \in [\beta_{\text{min}},\infty)^d$ controls the range to which the measurements
are restricted to, and $\hat{\bm\mu}\in \R^d$ is the global mean of the data, considered as a fixed constant. In this work, we let $\beta_{\text{min}}=1$.
Additionally, we consider a ratio of winsorization to apply to each predictor variable, controlled by an unknown parameter vector $\bm \alpha \in [0,1]^d$, combined with the smoothed winsorization operator \eqref{eq:adaptive_windsorization} via a residual connection. This gives the following adaptive outlier mitigation operation for an input time series $\bfx_t^{(i)} \in \R^d$:
\begin{equation}
    \mathbf{h}_1 \left(\bfx_t^{(i)}\right) = \bm\alpha \odot \breve{\bfx}_t^{(i)} + \left(\boldsymbol 1_d -\bm\alpha \right) \odot \bfx_t^{(i)},
\end{equation}
%where the ratio of winsorization to apply to each predictor variable is controlled by
%the unknown parameter $\bm \alpha \in [0,1]^d$ and the range to which the measurements
%are restricted to is controlled by a parameter$\bm \beta \in [\beta_{\text{min}},\infty)^d$.
where $\boldsymbol 1_d$ is a $d$-dimensional vector of ones.
Both $\bm\alpha$ and $\bm\beta$ are considered as unknown parameters as part of the %training objective. % automatically adapted to the training 
full objective function optimised during training.

%For input vector $\bfx_t^{(i)} \in \R^d$, 
%EDAIN applies the adaptive outlier removal operation with:
%\begin{equation}
%\begin{aligned}\label{eq:adaptive-outlier-removal}
%    \mathbf{h}_1&\left(\bfx_t^{(i)}\right)=\\
%    &\bm\alpha \odot \underbrace{\left(\bm\beta \odot
%        \tanh\left\{\left(\bfx_t^{(i)}-\hat{\bm\mu}\right) \oslash \bm\beta  \right\}+\hat{\bm\mu}
%\right)}_{\text{smooth adaptive centered winsorization}}
%     \\
%    +&\underbrace{\left(1-\bm\alpha \right) \odot \bfx}_{\text{residual connection}}.
%\end{aligned}
%\end{equation}

%The $\hat{\bm\mu}\in \R^d$ vector in equation
%\eqref{eq:adaptive-outlier-removal} is the global mean of
%the data, considered as a fixed quantity.

\paragraph{Shift and scale sublayers.}

The adaptive shift and scale layer, combined, perform the operation\begin{equation}\label{eq:adaptive-scale-shift}
    \mathbf{h}_3\left\{\mathbf{h}_2\left(\bfx_t^{(i)}\right)\right\}=(\bfx_t^{(i)} - \bm
    m) \oslash \bm s,
\end{equation}
where the unknown parameters are $\bm m \in \R^d$ and $\bm s \in (0,\infty)^d$.
Note that the EDAIN scale and shift sublayers generalise $z$-score scaling, which does not treat $\bm m$ and $\bm s$ as unknown parameters, but it sets them to the mean and standard deviation instead:
\begin{equation}
    \bm m=\frac{1}{N T}  \sum^{N}_{i=1} \sum^{T}_{t=1} \bfx_{t}^{(i)},
\end{equation}
and
\begin{equation}
    \bm s=\sqrt{\frac{1}{N T} \sum^{N}_{i=1} \sum^{T}_{t=1} \left(\bfx_{t}^{(i)}- \bm m\right)^2},
\end{equation}
where the power operations are applied element-wise.

\paragraph{Power transform sublayer.}

Many real-world datasets exhibit significant skewness, which is often corrected using power
transformations \citep{skewed_data},
%The most common transformation is 
such as the commonly used Box-Cox transformation \citep{boxcox}. One of the main limitations of the Box-Cox transformation is that it is only valid for positive values. %, which is a significant limitation for applications to many real-world datasets. 
A more general alternative which is more widely applicable is the Yeo-Johnson (YJ) transform \citep{yeoJohnson}:
\begin{equation}\label{eq:yeo-johnson}
    f_{\textrm{YJ}}^\lambda(x)= \left\{
        \begin{array}{ll}
            \frac{(x+1)^{\lambda}-1}{\lambda} & \textrm{if } \lambda \neq 0,\ x \geq 0, \\
            \log(x + 1) & \textrm{if } \lambda = 0,\ x \geq 0, \\
            \frac{(1-x)^{2-\lambda}-1}{\lambda-2} & \textrm{if } \lambda \neq 2,\ x < 0, \\
            -\log(1-x) & \textrm{if } \lambda=2,\ x < 0.
        \end{array}
    \right.
\end{equation}
The transformation $f_{\textrm{YJ}}^\lambda$ only has one unknown parameter, $\lambda\in\mathbb R$, and
it can be applied to any $x \in \R$, not just positive values \citep{yeoJohnson}.
The power transform sublayer of EDAIN simply applies the transformation in
Equation \eqref{eq:yeo-johnson} along each dimension of the input time series $\bfX^{(i)}$.
That is, for each
$i=1,\dots,N$ and $t=1,\dots,T$, %we define
the sublayer outputs
\begin{equation}
\mathbf{h}_4\left(\bfx_t^{(i)}\right) = 
\left[
f^{\lambda_1}_{\textrm{YJ}}\left(x_{t,1}^{(i)}\right),
\dots,
f^{\lambda_d}_{\textrm{YJ}}\left(x_{t,d}^{(i)}\right)\right], %\;\;\; j=1,\dots,d,
\end{equation}
where the unknown quantities to be optimised are the power parameters $\bm\lambda=(\lambda_1,\dots,\lambda_d) \in \R^d$.

\subsection{Global- and local-aware normalization}%
\label{sub:global_and_local}

\begin{figure}[t]
    \centering
    \includegraphics[width=0.48\textwidth]{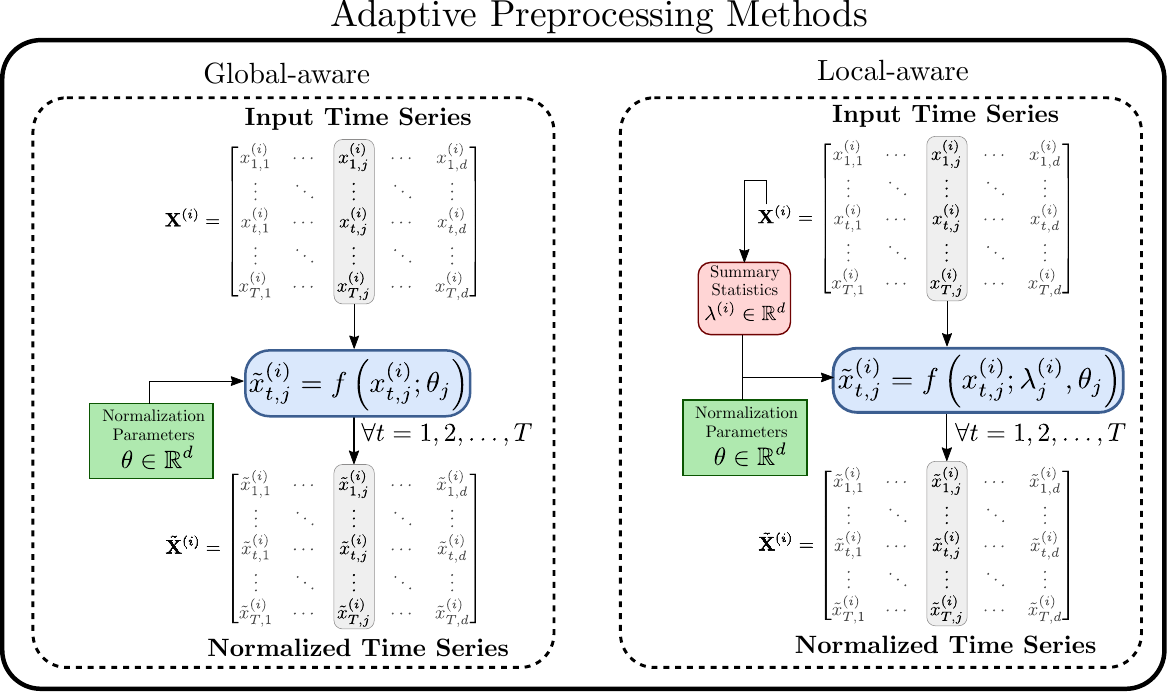}%
    \caption{
    Visual comparison of the local- and global-aware versions of
    adaptive preprocessing schemes.}%
    \label{fig:mode-vs-global}
\end{figure}

% TODO: introduce local and global-aware versions
For highly multi-modal and non-stationary time series data,
\cite{dain,rdain} and \cite{bin} observed significant performance improvements
when using local-aware preprocessing methods, as these
allow forming a unimodal representation space from predictor variables
with multi-modal distributions.
Therefore, we also propose a \textit{local-aware} version of the EDAIN layer in addition to the \textit{global-aware} version we presented 
earlier.
% Shift and scale operation for local-aware mode
In the local-aware version of EDAIN, the shift and scale operations also
depend on a summary representation of the current time series
$\bfX^{(i)}$ to be preprocessed:
%\begin{equation}
\begin{equation}
    \mathbf{h}_3\left\{\mathbf{h}_2\left(\bfx_t^{(i)}\right)\right\} =
    \left\{\bfx_t^{(i)} - \left(\bm m \odot \bm\mu_{\bfx}^{(i)}\right)\right\} \oslash \left(\bm s \odot \bm\sigma_{\bfx}^{(i)}\right).
\end{equation}
The summary representations $\bm\mu_{\bfx}^{(i)},\bm\sigma_{\bfx}^{(i)} \in \R^d$,
%which, as shown in \autoref{fig:mode-vs-global},  
are computed through a reduction
along the temporal dimension of each time series $\bfX^{(i)}$ (\textit{cf.} \autoref{fig:mode-vs-global}):
\begin{equation}
    \bm\mu_\bfx^{(i)}=\frac{1}{T} \sum^{T}_{t=1} \bfx^{(i)}_t,\ 
    \bm\sigma_\bfx^{(i)}=\sqrt{\frac{1}{T}  \sum^{T}_{t=1} \left(\bfx^{(i)}_t- \bm\mu_\bfx^{(i)} \right)^2}, \label{eq:sigma-summary-repr}
\end{equation}
where the power operations are applied element-wise in \eqref{eq:sigma-summary-repr}.
The outlier mitigation and power transform sublayers remain the same
for the local-aware version, except the $\hat{\bm\mu}$ statistic in
Equation \eqref{eq:adaptive_windsorization} is no longer fixed, but
rather the mean of the input time series:
\begin{equation}
    \hat{\bm\mu}^{(i)}=\frac{1}{T} \sum_{t=1}^T \bfx_{t}^{(i)}.
\end{equation}

% Local-aware vs. global-aware modes diagram
% TODO: change bottom border to by fully-filled

% DONE: introduce global-aware normalization
%Not all real-world datasets are multi-modal. Instead, they may contain measurements from
%a predictor variables with one dominant mode, but whose distribution is still significantly
%skewed and contains outliers. This is often the case if all the measurements originate from
%the same data generation mechanism, as opposed to multiple in multi-modal data.

\paragraph{Order preservation.}
Adaptive local-aware
preprocessing methods work well on multi-modal data such as financial
forecasting datasets \citep{dain,rdain,bin}.
%However, not all real-world data exhibits multi-modality.
However, as we will experimentally demonstrate in the next section,
local-aware preprocessing methods might perform worse than conventional
methods such as $z$-score normalization when applied to
unimodal data. 
This is because local-aware methods are not \textit{order preserving}, since the shift and scale
amount is different for each time series.
In many applications in which the features have a natural and meaningful ordering, 
%(DAIN, RDAIN, BIN, and local-aware EDAIN) because the shift and scale
%amount is different for each time series.
%This could be explained by considering the following
%fact:
%, in many applications 
it would be desirable to have
%In such a case, we want the normalization transformation to preserve the relative ordering
%between measurements across the sample indices,
%that is
%, we have the implication:
\begin{equation}\label{eq:order-preserving}
    \tilde{x}^{(i)}_{t,k} < \tilde{x}^{(j)}_{t,k}\ \mathrm{if}\ x^{(i)}_{t,k} < x^{(j)}_{t,k},%\implies
\end{equation}
for all $i,j \in \{1,\dots,N\}$, $t=1,\dots,T$ and $k=1,\dots,d$, where $\tilde{\bfX}^{(i)}$ denotes the output of a transformation with
$\bfX^{(i)}$ as input.
%where 
This property does not necessarily hold for the local-aware methods
(DAIN, RDAIN, BIN, and local-aware EDAIN). % because the shift and scale
%amount is different for each time series.
For unimodal features, the qualitative interpretation of the
predictor variables might dictate that property \eqref{eq:order-preserving} should be maintained (for example, consider the case of credit scores for a default prediction application, \textit{cf.} \autoref{sec:defpred}).

As a solution, the proposed \textit{global-aware} version of the proposed
EDAIN layer does not use any time series specific summary statistics, which makes
each of the four sublayers monotonically non-decreasing.
%It can be shown that t
This ensures that property \eqref{eq:order-preserving}
is maintained by the global-aware EDAIN transformation, providing additional flexibility for applications on real world data where ordering within features should be preserved. %, which might
%explain why it demonstrates superior performance on two of the datasets
%considered in the next section.

% DONE: Describe briefly how the parameters are updated, and refer to DAIN wrt. variable base learning rates
\subsection{Optimising the EDAIN layer}

The output of the proposed EDAIN layer is obtained by feeding the input time series
$\bfX^{(i)}$ through the four sublayers in a feed-forward fashion, as shown in
\autoref{fig:edain-arch}. The output is then fed to the deep neural network used for
the task at hand. Letting $\mathbf{W}$ denote the weights of the deep neural network model,
the weights of both the deep model and the EDAIN layer are simultanously optimised in an
end-to-end manner using stochastic gradient descent, with the update equation:
\begin{multline}
    \Delta \left( \bm{\alpha}, \bm{\beta}, \bm{m}, \bm{s}, \bm{\lambda}, \mathbf{W}\right)
    = \\
    -\eta\bigg(
        \eta_1 \frac{\partial \mathcal{L}}{\partial \bm\alpha},
        \eta_1 \frac{\partial \mathcal{L}}{\partial \bm\beta},
        \eta_2 \frac{\partial \mathcal{L}}{\partial \bm m},
        \eta_3 \frac{\partial \mathcal{L}}{\partial \bm s},
        \eta_4 \frac{\partial \mathcal{L}}{\partial \bm{\lambda}},
        \frac{\partial \mathcal{L}}{\partial \mathbf{W}}
    \bigg),
\end{multline}
where $\eta \in \R_+$ is the base learning rate, whereas $\eta_1,\dots,\eta_4 \in \R_+$ correspond to sublayer-specific corrections to the global learning rate $\eta$.
As \cite{dain} observed when training their DAIN layer, the gradients of the unknown
parameters for the different sublayers might have vastly different magnitudes,
which prevents a smooth convergence of the preprocessing layer. Therefore,
they proposed using separate learning rates for the different sublayers.
We therefore introduce corrections $\eta_\ell,\ \ell=\{1,2,3,4\}$ as additional hyperparameters that modify the learning rates for each of the four 
different EDAIN sublayers.

Furthermore, note that computing the fixed constant $\hat{\bm\mu}$ in the outlier mitigation sublayer \eqref{eq:adaptive_windsorization} would require a sweep on the entire dataset before training the EDAIN-augmented neural network architecture, which could be computationally extremely expensive. As a solution to circumvent this issue, we propose to calculate $\hat{\bm\mu}$ iteratively during training, updating it using
a cumulative moving average estimate at each forward pass of the
sublayer. We provide more details on this in \autoref{sec:mean_estimation_sup}.

% DONE: brief explanation of EDAIN-KL, emphasizing that it is also a novel invention, also motivate it
\subsection{EDAIN-KL}

In addition to the EDAIN layer, we also propose another novel preprocessing method, named
EDAIN-KL (Extended Deep Adaptive Input Normalization, optimised with Kullback-Leibler 
divergence). This approach uses a similar neural layer architecture as the EDAIN method,
but modifies it to ensure the transformation is invertible. Its unknown parameters are
then optimised with an approach inspired by normalizing flows \citep[see, for example,][]{normalizing_flows}. 

The EDAIN-KL layer is used to
transform a Gaussian base distribution $\mathbf{Z}\sim\mathcal{N}(\bm{0}, I_{dT})$ via a composite function $\mathbf{g}_{\boldsymbol{\theta}}=\mathbf{h}_1^{-1} \circ \mathbf{h}_2^{-1} \circ \mathbf{h}_3^{-1} \circ \mathbf{h}_4^{-1}$ comprised of the inverses of the operations in the EDAIN sublayers, applied sequentially with parameter $\boldsymbol{\theta} = (\boldsymbol{\alpha},\boldsymbol{\beta},\boldsymbol{m},\boldsymbol{s},\boldsymbol{\lambda})$. The parameter $\boldsymbol{\theta}$ is chosen to minimize the KL-divergence between the
resulting distribution $\mathbf{g}_{\boldsymbol{\theta}}(\mathbf Z)$ and the empirical distribution
of the dataset $\mathcal D$: 
\begin{equation}
\hat{\boldsymbol{\theta}}=\argmin_{\boldsymbol\theta}\mathrm{KL}\left\{\mathcal D\ \vert\vert\ \mathbf{g}_{\boldsymbol{\theta}}(\mathbf Z)\right\}.
\end{equation}
Note that we apply all the operations in reverse order, compared to the EDAIN layer, because we use $\mathbf{g}_{\boldsymbol{\theta}}$ to transform a base distribution $\mathbf Z$ into a distribution that resembles the training dataset $\mathcal D$. %After obtaining $\hat{\boldsymbol{\theta}}$, 
To normalize the dataset after fitting the EDAIN-KL layer, we apply
\begin{equation}
    \mathbf{g}^{-1}_{\hat{\boldsymbol{\theta}}} = \mathbf{h}_4 \circ \mathbf{h}_3 \circ \mathbf{h}_2 \circ \mathbf{h}_1
\end{equation}
to each $\bfX^{(i)} \in \R^{d \times T}$, similarly to the EDAIN layer.
%As such, the optimisation objective
%becomes making the data as normally distributed as possible.
The main advantage of the EDAIN-KL approach over standard EDAIN is that it allows training in
an unsupervised fashion, separate from
the deep model. This enables its usage for preprocessing data in a wider set of tasks,
including %preprocessing data for 
non-deep-neural-network models.
An exhaustive description of the EDAIN-KL method
is provided in \autoref{sec:edain_kl_sub}.

\section{EXPERIMENTAL EVALUATION}%
\label{sec:experimental_evaluation}

% Introduce this section
For evaluating the proposed EDAIN layer we consider a synthetic dataset, a large-scale default prediction dataset, and a large-scale financial forecasting dataset.
% Explain the methods we compare to
We compare the two versions of the EDAIN layer (global-aware and local-aware) and the EDAIN-KL layer to %$z$-score normalization, %, a very common
%static preprocessing method, 
the DAIN \citep{dain} layer, to the 
BIN \citep{bin} layer and to the KDIT method.
We also consider a statistical baseline consisting
of different combinations of $z$-score scaling,
winsorization and a Yeo-Johnson power
transformation.
% Explain briefly the RNN models used, incl. note about categorical features
For all experiments, we use a recurrent neural network (RNN)
model composed of gated recurrent unit (GRU) layers,
followed by a classifier head with fully connected
layers. %For the default prediction dataset which contains some
Categorical features, when present, %these 
are passed through an embedding layer,
whose output is combined with the output of the GRU layers and
then fed to the classifier head. Full details on the model
architectures, optimization procedures, including learning rates and
number of epochs, can be found in \autoref{sec:exp_eval_details} and in the code repository associated with
this work.

\subsection{Synthetic Datasets}%
\label{sub:synth_dat}

% TODO: ...
Before considering real-world data, we evaluate our method on synthetic
data, where we have full control over the data generating process.
% Briefly explain novel data generation mechanism in 2-3 sentences
To do this, we develop a synthetic time series data generation algorithm which allows
specifying arbitrary unnormalized probability density functions (PDFs)
for each of the $d$ predictor variables. It then generates $N$ time series of the form $\bfX^{(i)} \in \R^{d \times T}$, along with $N$ binary response variables ${y}^{(i)} \in \{0,1\}$. We present a detailed description of the algorithm
in \autoref{sec:synth-data-appendix}.

\begin{figure}[t]
    \centering
    \includegraphics[width=0.48\textwidth,trim={0 7 0 7pt},clip]{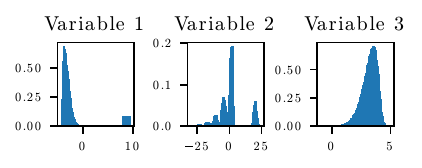}%
    \caption{Histogram across timesteps $t=1,\dots,T$ of the $d=3$ predictor variables from the synthetic data.}%
    \label{fig:synth-eda}
\end{figure}

% Evaluation methodology: Metrics, number of datasets generated, N, d, T
For our experiments, we generated $N_\mathcal{D}=100$ datasets, each
with $N=50\,000$ time series of length $T=10$ and dimensionality
$d=3$. The three predictor variables were configured to be distributed as follows:
\begin{align}
    f_1(x)=&\ 10 \cdot \Phi_{\mathcal{N}}\left\{10\left(x+4\right)
    \right\} \cdot p_{\mathcal{N}}(x+4) \nonumber \\ &+  \mathbb{I}_{(8,9.5)}(x) \cdot e^{x-8}/10, \label{eq:synth_pdf1}\\
    f_2(x)=& \left\{\begin{array}{ll}
        20 \cdot p_{\mathcal{N}}(x-20), & \textrm{ if } x > \pi, \\
            e^{x/6}\cdot\left\{10\sin(x)+10 \right\}, & \textrm{ if } x \leq \pi,
    \end{array}\right. \label{eq:synth_pdf2}\\
        f_3(x)=&\
        2 \cdot \Phi_{\mathcal{N}}\left\{-4 ( x-4)\right\} \cdot p_{\mathcal{N}}(x-4), \label{eq:synth_pdf3}
\end{align}
where $p_\mathcal{N}(\cdot)$ and $\Phi_\mathcal{N}(\cdot)$ %respectively 
denote the
PDF and cumulative distribution function (CDF) of the standard normal
distribution, and $\mathbb{I}_\mathcal{A}(\cdot)$ is the indicator function on the set $\mathcal{A}$. 
Samples from the dataset are visualised in \autoref{fig:synth-eda}.
We train and evaluate a RNN model with the architecture described
earlier on each of the $N_{\mathcal{D}}$ datasets using a 80\%-20\%
train-validation split. Our results are presented in \autoref{tab:synth-results}, where the binary cross-entropy (BCE) loss 
and the %binary 
accuracy on the validation set are used as evaluation metrics.

% Draw conclusions
From our experiments on the synthetic datasets, we observe that the model performance is more unstable when no preprocessing is applied, as seen from the increased variance in \autoref{tab:synth-results}.
We also observe that $z$-score normalization only gives minor performance improvements when compared to no preprocessing, aside from reducing the variance.
%% Start explanation of CDF inversion %%
As we have perfect information about the underlying data generation
mechanism from Equations \eqref{eq:synth_pdf1}, \eqref{eq:synth_pdf2} and \eqref{eq:synth_pdf3}, we also compared our methods to what we refer
to as \textit{CDF inversion}, where each observation is transformed by the CDF
of its corresponding distribution, and then transformed via
$\Phi_{\mathcal{N}}^{-1}(\cdot)$, giving predictor variables with
standard normal distributions. 
We also apply this method to the real-world datasets in
\autoref{sec:defpred} and \autoref{sub:lob_data}, but since the true PDFs are unknown in those settings, we estimate the CDFs using
quantiles from the distribution of the training data.
The CDF inversion method resembles KDIT for low
$\alpha$, since KDIT becomes a quantile transformation in
the limiting case $\alpha \rightarrow 0$ \citep{kdit}. 
The main difference is that the KDIT method does not involve
the Gaussianization step, implying that the preprocessing matches the data generation mechanism, representing a gold-standard for this example. As discussed in more details in
\autoref{sec:synth-data-appendix}, the synthetic responses
are generated from a linear combination of uniform random
variables, so the KDIT gives the model a stronger signal
than that of the CDF inversion method.
%% End explanation of CDF inversion %%
Out of all the methods not exploiting the
mechanics of the underlying
data generation mechanism, the global-aware version of EDAIN demonstrates superior performance. It also almost performs as well as CDF inversion, which is able to perfectly normalize each predictor variable via its data generation mechanism.
Finally, we observe that the local-aware methods perform even worse than no preprocessing: this might be due to the ordering not being preserved,
as discussed in \autoref{sub:global_and_local}.

\begin{table}[t]
\resizebox{0.48\textwidth}{!}{%
\begin{tabular}{l|ccc}
\toprule
Preprocessing method                    & BCE loss                  & Binary accuracy (\%)             \\
\midrule
No preprocessing & $0.1900 \pm 0.0036$ & $0.9168 \pm 0.0017$ \\
$z$-score & $0.1871 \pm 0.0011$ & $0.9176 \pm 0.0007$ \\
$z$-score + YJ & $0.1789 \pm 0.0009$ & $0.9211 \pm 0.0006$ \\
Winsorize + $z$-score & $0.1876 \pm 0.0010$ & $0.9172 \pm 0.0006$ \\
Winsorize + $z$-score + YJ & $0.1788 \pm 0.0009$ & $0.9211 \pm 0.0006$ \\
\underline{CDF inversion} & $\underline{0.1627 \pm 0.0009}$ & $\underline{0.9289 \pm 0.0006}$ \\
BIN & $0.2191 \pm 0.0010$ & $0.9036 \pm 0.0006$ \\
DAIN & $0.2153 \pm 0.0015$ & $0.9048 \pm 0.0008$ \\
EDAIN (local-ware) & $0.2099 \pm 0.0010$ & $0.9071 \pm 0.0006$ \\
\textbf{EDAIN (global-aware)} & $\mathbf{0.1636 \pm 0.0009}$ & $\mathbf{0.9283 \pm 0.0005}$ \\
EDAIN-KL & $0.1760 \pm 0.0009$ & $0.9224 \pm 0.0006$ \\
\underline{KDIT} ($\alpha=0.1$) & $\underline{0.1532 \pm 0.0011}$ & $\underline{0.9329 \pm 0.0006}$ \\
%KDIT ($\alpha=1$) & $0.2132 \pm 0.0017$ & $0.9059 \pm 0.0009$ \\
%KDIT ($\alpha=10$) & $0.2492 \pm 0.0025$ & $0.8901 \pm 0.0012$ \\
%KDIT ($\alpha=100$) & $0.2546 \pm 0.0031$ & $0.8876 \pm 0.0015$ \\
\bottomrule
\end{tabular}}%
\caption{Experimental results on synthetic data, with 95\% normal confidence intervals 
$\mu\pm1.96\sigma/\sqrt{K}$
calculated across $K=100$ datasets. The gold-standard CDF-based transformations are underlined.}%
\label{tab:synth-results}
\end{table}

\subsection{Default Prediction Dataset}%
\label{sec:defpred}

% DONE: Introduce and describe the dataset
The first real-world dataset we %will test our proposed processing methods on 
consider is
the publicly available default prediction dataset published by American Express
\citep{amex-data}, which contains data from $N=458\,913$ credit card customers.
For each customer, a vector of $d=188$ aggregated profile features has been
recorded at $T=13$ different credit card statement dates, producing a multivariate
time series of the form $\bfX^{(i)}\in \R^{d \times T},\ i=1,2,\dots,N$. Given $\bfX^{(i)}$, the task is to predict
a binary label $y^{(i)} \in \{0,1\}$ indicating whether the $i$-th customer 
defaulted at any point within 18 months after the last observed data point in the time series. %or not, where 
A default event is defined as not paying back the credit card
balance amount within 120 days after the latest %credit card 
statement date
\citep{amex-data}. 
Out of the $d=188$ features, only the 177 numerical variables are preprocessed.

% In the dataset provided by
% \cite{amex-data}, the non-default records have been \textit{down-sampled} such
% that the final default rate in the dataset is about 25\%.  Additionally, the
% feature names have all been anonymized, but they can be categorised into the
% five categories: delinquency variables, spend variables, payment variables,
% balance variables, and risk variables \citep{amex-data}.  

% DONE: Explain evaluation methodology
To evaluate the different preprocessing methods, we perform 5-fold cross validation, which produces evaluation metrics for five different 20\%
validation splits.
%Extensive details about all  the hyperparameters used in our experiments
%are contained in the supplementary material (TODO: ref)?
%TODO: details on CDF inversion in this case
%We also compare the methods to a CDF inversion technique (\textit{cf.} \autoref{sub:synth_dat}). Since the true PDFs is unknown, we estimate the CDFs using
%quantiles from the distribution of the training data.
%
% DONE: explaining the evaluation metrics
The evaluation metrics we consider are the validation BCE loss and a metric that was proposed by \cite{amex-data} for use with this dataset, and we refer to it as the
\textit{Amex metric}.
This metric is calculated as $M=0.5 \cdot (G+D)$, where $D$ is the default rate
captured at 4\% (corresponding to the proportion of positive labels captured within the
highest-ranked 4\% of the model predictions) and $G$ is the normalized
Gini coefficient %, of which a full definition is provided in the supplementary material
%\citep[see, for example,][]{gini}.
(see \autoref{sec:amex_metric_sub} and, for example, \citeauthor{gini}, \citeyear{gini}).

\begin{table}[t]
\resizebox{0.48\textwidth}{!}{%
\begin{tabular}{l|cc}
\toprule
Preprocessing method & BCE loss & AMEX metric \\
\midrule
No preprocessing & $0.3242 \pm 0.0066$ & $0.6430 \pm 0.0087$ \\
$z$-score & $0.2213 \pm 0.0017$ & $0.7872 \pm 0.0030$ \\
Winsorize + $z$-score & $0.2217 \pm 0.0018$ & $0.7867 \pm 0.0025$ \\
$z$-score + YJ & $0.2224 \pm 0.0014$ & $0.7846 \pm 0.0022$ \\
Winsorize + $z$-score + YJ & $0.2926 \pm 0.1229$ & $0.6321 \pm 0.2679$ \\
CDF inversion & $0.2215 \pm 0.0018$ & $0.7861 \pm 0.0032$ \\
EDAIN-KL & $0.2218 \pm 0.0018$ & $0.7858 \pm 0.0027$ \\
EDAIN (local-aware) & $0.2245 \pm 0.0015$ & $0.7813 \pm 0.0025$ \\
EDAIN (global-aware) & $0.2199 \pm 0.0015$ & $0.7890 \pm 0.0035$ \\
DAIN & $0.2224 \pm 0.0016$ & $0.7847 \pm 0.0024$ \\
BIN & $0.2237 \pm 0.0017$ & $0.7829 \pm 0.0029$ \\
%KDIT ($\alpha=0.1$) & $0.2277 \pm 0.0012$ & $0.7751 \pm 0.0026$ \\
KDIT ($\alpha=1$) & $0.2258 \pm 0.0014$ & $0.7791 \pm 0.0024$ \\
%KDIT ($\alpha=10$) & $0.2272 \pm 0.0013$ & $0.7782 \pm 0.0020$ \\
%KDIT ($\alpha=100$) & $0.2290 \pm 0.0014$ & $0.7750 \pm 0.0019$ \\

\bottomrule
\end{tabular}}%
\caption{Experimental results on the default prediction dataset, with 95\% normal asymptotic confidence intervals $\mu\pm1.96\sigma/\sqrt{K}$ calculated across $K=5$ folds.}%
\label{tab:amex-results}
\end{table}

\begin{figure}[t]
    \centering
    \includegraphics[width=0.48\textwidth]{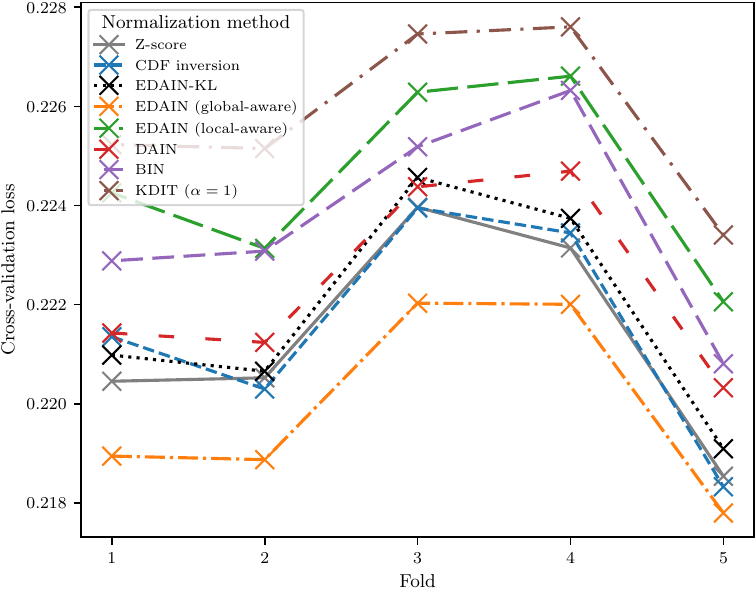}
    \caption{BCE cross-validation loss across different folds in the Amex default prediction dataset.}
    \label{fig:amex-fold-breakdown}
\end{figure}

% DONE: discuss the results
Our results are reported in Table \ref{tab:amex-results}. From the table, it can be inferred that neglecting the preprocessing step
deteriorates the performance significantly. Moreover, we observe that the local-aware methods
(local-aware EDAIN, BIN, and DAIN) all perform worse than $z$-score normalization. This might be
because the data in the dafault prediction dataset is mostly distributed around one
central mode, and the local-aware methods discard this information in favour of forming
a common representation space. Another possible reason is that the local-aware methods
do not preserve the relative ordering between data points $\bfX^{(i)}$ as per
Equation \eqref{eq:order-preserving}, which might be detrimental for these types of datasets. As discussed in the previous section, predictor variables such as credit scores may be present in the dataset, which should only be preprocessed via 
monotonic transformations.
From the results presented in \autoref{tab:amex-results}, we can also conclude that the
proposed global-aware version of the EDAIN layer shows superior average performance when compared 
to all %the local-aware preprocessing methods and conventional 
alternative preprocessing methods.
Additionally, it must be remarked that most of the variance in the methods' performance arise from the folds themselves,
as seen in \autoref{fig:amex-fold-breakdown}, and  the global-aware EDAIN method consistently shows
superior performance across all cross-validation folds when compared to the other methods. This was further confirmed via a paired sign test between EDAIN and BIN (the second best performing method), which returned $p$-value $0.015$, indicating a significant difference in the performance of the two preprocessing methods.

\subsection{Financial Forecasting Dataset}%
\label{sub:lob_data}

% DONE: Introduce and describing the dataset
For evaluating the proposed methods, we also considered a limit order book (LOB) time series dataset \citep[FI-2010 LOB][]{lob-data}, used as a benchmark dataset in \cite{dain} and \cite{bin}.
The data was collected across 10 business days in June 2010
from five Finnish companies \citep{lob-data}, and 
it was cleaned and features were extracted based on the pipeline
%proposed by 
of \cite{lob_preprocess}.
This resulted in $N=453\,975$ vectors of dimensionality $d=144$.
The task is predicting whether the mid price will increase, decrease or remain
stationary with a prediction horizon of $H=10$ timesteps, where a stock is labelled as 
stationary if the mid price changes by less than 0.01\%.
More details on the FI-2010 benchmark dataset can be found in \cite{lob-data}.

% DONE: Explainaing the cross-validation scheme
For training, we use the anchored cross-validation scheme of \cite{lob-data}.
%We use the first day of data to train the model, then evaluate it on the second
%day of data. Afterwards, the first two days of data are used for training, and the third day
%for evaluation. This is repeated until we use the first 9 days of data for training and the 10th day for 
%evaluation. 
We sequentially increase the size of the training set, starting with a single day and extending it to nine days, while reserving the subsequent day for evaluation in each iteration, obtaining nine different evaluation folds.
%Full details on the optimization
%procedure, including learning rates and number of epochs can be found in
%the supplementary material (TODO: reference) and code repository associated with this work.
%
% Optional: Mention that data was sampled with inverse proportion to class frequency to ensure balance?
%
% DONE: Explaining the metrics
For evaluating the model performance, %on the FI-2010 LOB dataset, 
we look at the Cohen's $\kappa$ \citep{kappa} and
macro-$F_1$ score, obtained by averaging class-specific $F_1$ scores across the three possible outcomes (increase, decrease or stationary).
%Let $\textrm{TP}_c$, $\textrm{FP}_c$, $\textrm{TN}_c$
%and $\textrm{FN}_c$ denote the true positives, false positives, true negatives and
%false negatives of class $c \in \mathcal{C}=\{\textrm{decrease}, \textrm{stationary}, \textrm{increase}\}$. The precision and recall of class $c$ is then calculated with
%$\textrm{prec}_c= {\textrm{TP}_c}/({\textrm{TP}_c+\textrm{FP}_c})$ and
%$\textrm{recall}_c={\textrm{TP}_c}/({\textrm{TP}_c+\textrm{FN}_c})$, respectively.
%The $F_1$-score for class $c$ is then calculated with:
%\begin{equation}
%    F_1^{(c)}=2\frac{\textrm{precision}_c  \cdot \textrm{recall}_c}{\textrm{precision}_c+\textrm{recall}_c}.
%\end{equation}
%These metrics are computed for each class $c \in \mathcal{C}$ separately, and then averaged
%to give the macro-$F_1$ score.
%We also use Cohen's $\kappa$ metric to measure the %agreement between the label predictions
%and the true label annotations, while accounting for agreement by chance.
%More details on the metric is provided by \cite{kappa}.

% In \autoref{tab:lob-results}, we report the mean and a 95\% asymptotic normal confidence interval
% ($\mu \pm 1.96 \sigma$) for each metric, based on the 9 different validation folds.

\begin{table}[t]
\resizebox{0.48\textwidth}{!}{%
\begin{tabular}{l|cc}
\toprule
Preprocessing method & Cohen's $\kappa$ & Macro-$F_1$-score \\
\midrule
No preprocessing & $0.0035 \pm 0.0016$ & $0.2859 \pm 0.0076$ \\
%Standard scaling & $0.2772 \pm 0.0183$ & $0.5047 \pm 0.0134$ \\
$z$-score & $0.2777 \pm 0.0185$ & $0.5052 \pm 0.0137$ \\
Winsorize + $z$-score & $0.2928 \pm 0.0205$ & $0.5166 \pm 0.0168$ \\
CDF inversion & $0.3618 \pm 0.0199$ & $0.5798 \pm 0.0124$ \\
BIN & $0.3670 \pm 0.0213$ & $0.5889 \pm 0.0160$ \\
DAIN & $0.3588 \pm 0.0169$ & $0.5776 \pm 0.0114$ \\
\textbf{EDAIN (local-aware)} & $\mathbf{0.3836 \pm 0.0185}$ & $\mathbf{0.5946 \pm 0.0144}$ \\
EDAIN (global-aware) & $0.2820 \pm 0.0235$ & $0.5111 \pm 0.0216$ \\
EDAIN-KL & $0.2870 \pm 0.0214$ & $0.5104 \pm 0.0173$ \\
KDIT ($\alpha=0.1$) & $0.2974 \pm 0.0225$ & $0.5260 \pm 0.0170$ \\
%KDIT ($\alpha=1$) & $0.2878 \pm 0.0248$ & $0.5155 \pm 0.0191$ \\
%KDIT ($\alpha=10$) & $0.2649 \pm 0.0269$ & $0.4952 \pm 0.0202$ \\
%KDIT ($\alpha=100$) & $0.2648 \pm 0.0263$ & $0.4946 \pm 0.0201$ \\
\bottomrule
\end{tabular}}%
\caption{Experimental results on the FI-2010 LOB dataset, with 95\% normal asymptotic confidence intervals $\mu\pm1.96\sigma/\sqrt{K}$ calculated across $K=9$ anchored folds.}
\label{tab:lob-results}
\end{table}

% TODO: comment and briefly discuss the results
We can draw several conclusions from the results reported in \autoref{tab:lob-results}.
Firstly, employing some form of preprocessing is essential, as not applying any preprocessing gives $\kappa$ values close to 0, which is what is expected to occur by random guessing.
Secondly, applying a local-aware normalization scheme (local-aware EDAIN, BIN and DAIN)
greatly improves performance when compared to conventional preprocessing methods such as
$z$-score scaling. %We also remark that these local-aware layers were all given
%the raw data as input, but were still able to learn how to appropriately normalize the
%time series for successful training of the deep learning model. 
Our third observation is that
the proposed local-aware EDAIN method outperforms both BIN and DAIN on average. Finally, we
note that most of the variability in the evaluation metrics arises from the folds themselves, similarly to the default prediction example (\textit{cf.} Figure~\ref{fig:amex-fold-breakdown}). Again, this was further assessed via a paired sign test between EDAIN and the second best performing method, resulting in a $p$-value $0.00097$, which confirms the significance of our results.

\subsection{Ablation Study}%
\label{sub:ablation}

To identify the effect of each of the four sublayers on the final predictive performance,
we conducted an ablation study. The results are reported in \autoref{tab:ablation}, where
OM and PT refer to the outlier mitigation and power transform sublayers, respectively. These
experiments were conducted on the default prediction dataset using the same model
architecture and evaluation procedure as described in \autoref{sec:defpred}. We observe that
just using adaptive shift and scale sublayers provides marginal gains over the $z$-score scaling
baseline. Meanwhile, introducing either the adaptive power transform or the
outlier mitigation sublayers reduces the loss to a greater extent, with the outlier mitigation
sublayer proving most effective.

\begin{table}[t]
\resizebox{0.48\textwidth}{!}{%
\begin{tabular}{l|cc}
\toprule
Preprocessing method & BCE loss & AMEX metric \\
\midrule
$z$-score & $0.2214 \pm 0.0018$ & $0.7873 \pm 0.0029$ \\
scale & $0.2214 \pm 0.0016$ & $0.7863 \pm 0.0029$ \\
shift & $0.2213 \pm 0.0016$ & $0.7871 \pm 0.0035$ \\
shift+scale & $0.2212 \pm 0.0018$ & $0.7872 \pm 0.0035$ \\
shift+scale+PT & $0.2207 \pm 0.0016$ & $0.7885 \pm 0.0026$ \\
OM+shift+scale & $0.2203 \pm 0.0015$ & $0.7884 \pm 0.0031$ \\
\textbf{OM+shift+scale+PT} & $\mathbf{0.2199 \pm 0.0015}$ & $\mathbf{0.7889 \pm 0.0031}$ \\
\bottomrule
\end{tabular}}%
\caption{Ablation study on the default prediction, with 95\% normal asymptotic confidence intervals $\mu\pm1.96\sigma/\sqrt{K}$ calculated across $K=5$ folds. In all but the first row, the global-aware
EDAIN layer is used with the corresponding subset of its sublayers enabled.
}
\label{tab:ablation}
\end{table}

\section{CONCLUSION}

% Summarise the EDAIN layer (TODO)
In this work, we proposed EDAIN, an adaptive data preprocessing layer which can be used to augment any deep neural network architecture that takes multivariate time series as input. EDAIN has four adaptive sublayers (outlier mitigation, shift, scale, and power transform). It also has two versions (local-aware and global-aware), which apply local or global transformations to each time series. Also, we proposed a computationally efficient variant of EDAIN, optimised via the Kullback-Leibler divergence, named EDAIN-KL.
% Summarise the results
The EDAIN layer's ability to increase the predictive
performance of the deep neural network was evaluated on a synthetic dataset,
a default prediction dataset, and a financial forecasting
dataset. On all datasets considered, either the local-aware or global-aware version of the proposed EDAIN layer consistently demonstrated superior performance.

% Discussion of results observed in section 4
In \autoref{sub:lob_data}, we observed that the local-aware preprocessing methods gave significantly better performance than the global-aware version of EDAIN and $z$-score normalization.
However, in \autoref{sub:synth_dat} and \autoref{sec:defpred} the opposite is
observed, with the global-aware version of EDAIN demonstrating superior performance and the local-aware methods being outperformed by $z$-score normalization.
% Explanation of why
We hypothesize these differences occur because local-aware methods do not preserve the relative ordering between observations,
%when they have a natural and meaningful ordering,
while the global-aware EDAIN method does. In the financial forecasting dataset, which is highly multi-modal, it appears that the observations'
feature values relative to their mode are more important than their absolute ordering. %On the other hand, the opposite happens for the default predition dataset, where features like credit score are likely to be used. %is more important in the
%multi-modal .
Such considerations should be taken into account
when deciding what adaptive preprocessing method is most suitable
for application on new data.

% Present future work 1: Other model archs
There are several directions for future work.
\cite{rdain} observed that the performance improvements from DAIN %compared to standard 
%scaling 
differed greatly between different deep neural network architectures.
Therefore, the effectiveness of 
EDAIN with other architectures could be further explored as only GRU-based RNNs were considered 
in this work.
% Future work 2: local-aware and global-aware together
Additionally, with the proposed EDAIN method, one has to manually
decide whether to apply local-aware or global-aware preprocessing.
This drawback could be eliminated by extending the proposed neural
layer to apply both schemes to each feature and adaptively learning
which version is most suitable. %for each individual feature.
% Future work 3: also treat missing values
Another common irregularity in real-world data is missing values
\citep{nawi,brits}. %, which the EDAIN layer is unable to treat.
%Therefore, %another possible research direction is extending
A possible direction would be to extend
EDAIN with an adaptive method for treating missing values
that makes minimal assumptions on the data generation mechanism
\citep[for example,][]{brits}. 

%\section*{DISCLAIMER}
%
%\textit{This disclaimer has been redacted to preserve the anonymity of the authors and their
%affiliations.} 

%%%%%%%%%%%%%%%%%%%%%%%% Bibliography %%%%%%%%%%%%%%%%%%%%%%

\bibliography{refs}

\appendix
\onecolumn
\thispagestyle{empty}
\aistatstitle{Supplementary Materials\\[.25em]
Extended Deep Adaptive Input Normalization for
    Preprocessing Time Series Data for
    Neural Network
}

\section{MEAN ESTIMATION IN THE OUTLIER MITIGATION LAYER}%
\label{sec:mean_estimation_sup}

The $\hat{\bm\mu}\in \R^d$ vector in Equation~\eqref{eq:adaptive_windsorization} is an estimate of 
the mean of the data, and it is used
to ensure the smoothed winsorization operation is centered. When using the local-aware version of
the EDAIN layer, it is the mean of the input time series data point:
\begin{equation}
    \hat{\bm\mu}^{(i)}=\frac{1}{T} \sum_{t=1}^T \bfx_{t}^{(i)}. %, \qquad k=1,\dots,d.
\end{equation}
In the global-aware version of EDAIN, we consider $\hat{\bm\mu} \in \R^d$ a fixed quantity, estimating the global mean of the data. 
As mentioned in the main paper, the $\hat{\bm\mu}$ estimate is
iteratively computed during training using a \textit{cumulative
moving average estimate} at each forward pass of the sublayer.
To do this, we keep track of the current estimated average at forward pass $n=1,2,3,\dots$, denoted $\hat{\bm\mu}^{(n)}$, and when we
process an input time series $\bfX^{(i)}$, we apply the update:
\begin{equation}
    \hat{\bm\mu}^{(n+1)}= \frac{nT\cdot\hat{\bm\mu}^{(n)}+\sum^{T}_{t=1} \bfx_t^{(i)}}{(n+1)T}.
\end{equation}
We also initialise
$\hat{\bm\mu}^{(0)}=\mathbf{0}$.

\section{EDAIN-KL}%
\label{sec:edain_kl_sub}

The {EDAIN-KL} layer has a very similar architecture to the
{EDAIN} layer (\textit{cf.} \autoref{sec:edain} in the main paper), but the unknown parameters are learned via a different optimization procedure. Unlike the {EDAIN} layer, the {EDAIN-KL} layer is not attached to the deep neural
network during training, but rather training in isolation
before training the neural network.
This is done by establishing an invertible mapping to transform a standard
normal distribution into a distribution that resembles
that of our training dataset. Then, after
the {EDAIN-KL} weights have been optimized, we use the layer in reverse to normalize the time series
from the training dataset before passing them to the neural network.

\subsection{Architecture}%
\label{sub:Architecture}

Aside from the outlier mitigation sublayer, the {EDAIN-KL} layer has
an identical architecture to the global-aware {EDAIN} layer.
The outlier mitigation transformation has been simplified to ensure its
inverse is analytic. Additionally, the layer no longer supports local-aware mode, as this
would make the inverse intractable. The {EDAIN-KL} transformations are:
\begin{align}
    \textrm{(outlier mitigation)} \qquad& \mathbf{h}_1\left(\bfx^{(i)}_t\right)=\bm\beta \odot \tanh\left\{(\bfx^{(i)}_t - \hat{\bm\mu}) \oslash \bm\beta \right\}+\hat{\bm\mu} \label{eq:kl1}\\
    \textrm{(shift)} \qquad& \mathbf{h}_2\left(\bfx^{(i)}_t\right)=\bfx^{(i)}_t- \bm m \label{eq:kl2}\\
    \textrm{(scale)} \qquad& \mathbf{h}_3\left(\bfx^{(i)}_t\right)=\bfx^{(i)}_t \oslash \bm s  \label{eq:kl3}\\
    \textrm{(power transform)} \qquad& \mathbf{h}_4\left(\bfx^{(i)}_t\right)=\left[
        f^{\lambda_1 }_{\textrm{YJ}}\left(x^{(i)}_{t,1}\right),
        % \quad f^{\lambda_2^ {(\textrm{YJ})}}_{\textrm{YJ}}\left(x^{(i)}_{t,1}\right)
        \quad 
        \cdots,
        \quad f^{\lambda_d}_{\textrm{YJ}}\left(x^{(i)}_{t,d}\right)
    \right], \label{eq:kl4}
\end{align}
where $f^{\lambda_i}_{\textrm{YJ}}(\cdot)$ is defined in the main paper in Equation~\eqref{eq:yeo-johnson}.

\subsection{Optimisation through Kullback-Leibler divergence}%
\label{sub:opt_with_kl}

The  optimisation approach used to train the {EDAIN-KL} method is inspired by normalizing flows \citep[see, for example,][]{normalizing_flows}.
Before describing the approach, we provide a brief overview of related notation and some
background on the concept behind normalizing flows. After this, we describe how the
{EDAIN-KL} layer itself can be treated as an invertible bijector to fit into the
normalizing flow framework. In doing so, we derive analytic and differentiable
expressions for certain terms related to the {EDAIN-KL} layer. 

\subsubsection{Brief background on normalizing flow}%
\label{ssub:Brief background on normalizing flow}

The idea behind normalizing flows is taking a simple random variable, such as a standard
Gaussian, and transforming it into a more complicated distribution, for example, one that
resembles the distribution of a given real-world dataset.
Consider a random variable $\bfZ \in \R^d$ with a known and
analytic expression for its {PDF} $p_\bfz: \R^d \rightarrow {\R}$. We refer to $\bfZ$ as
the \textit{base distribution}. We then define
a parametrised invertible function
$\mathbf{g}_{\bm \theta}:\R^d \rightarrow {\R^d}$, also known as a \textit{bijector},
and use this to transform the base distribution into a new
 probability distribution: $\bfY=\mathbf{g}_{\bm\theta}(\bfZ)$.
By increasing the complexity of the bijector $\mathbf{g}_{\bm\theta}$ (for example, by using
a deep neural network), the transformed distribution $\bfY$ can grow arbitrarily complex as well.
The {PDF} of the transformed distribution can then be computed using the change of variable
formula \citep{normalizing_flows}, where
\begin{equation}
    p_\bfY(\bfy)=p_\bfZ(\mathbf{g}_{\bm\theta}^{-1}\left(\bfy\right))\cdot \left|\det \mathbf{J}_{\bfY \rightarrow {\bfZ}}\left(\bfy \right) \right| 
                 = p_\bfZ(\mathbf{g}_{\bm\theta}^{-1}\left(\bfy\right))\cdot \left|\det \mathbf{J}_{\bfZ \rightarrow {\bfY}}\left( \mathbf{g}^{-1}_{\bm\theta}(\bfy) \right) \right|^{-1}\label{eq:sj98days9dohao},
\end{equation}
and where $\mathbf{J}_{\bfZ \rightarrow {\bfY}}$ is the Jacobian matrix for the \textit{forward mapping}
$\mathbf{g}_{\bm\theta} : \bfz \mapsto \bfy$. 
Taking logs on both sides of Equation~\eqref{eq:sj98days9dohao}, it follows that
\begin{equation}\label{eq:logDetJac}
    \log p_\bfY(\bfy)= \log p_\bfZ(\mathbf{g}_{\bm\theta}^{-1}\left(\bfy\right)) - \log \left|\det \mathbf{J}_{\bfZ \rightarrow {\bfY}}\left(\mathbf{g}_{\bm\theta}^{-1}\left(\bfy\right) \right) \right|.
\end{equation}

One common application of normalizing flows is density estimation \citep{normalizing_flows}.
Given a dataset $\mathcal{D}=\{\bfy^{(i)}\}_{i=1}^N$ with samples from some
unknown, complicated distribution, we want to estimate its {PDF}.
This can be done with likelihood-based estimation, where we assume the data points
$\bfy^{(1)}, \bfy^{(2)},\dots,\bfy^{(N)}$ come from, say,
the parametrised distribution $\bfY=\mathbf{g}_{\bm \theta}(\bfZ)$ and
we optimise $\bm\theta$ to maximise the data log-likelihood,
\begin{equation}
    \log p(\mathcal{D}| \bm\theta)
     = \sum_{i=1}^N \log p_\bfY(\bfy^{(i)}| \bm\theta ) 
    \stackrel{\eqref{eq:logDetJac}}{=} \sum^{N}_{i=1}
    \log p_\bfZ\left(\mathbf{g}_{\bm\theta}^{-1}\left(\bfy^{(i)}\right)\right) - \log \left|\det \mathbf{J}_{\bfZ \rightarrow {\bfY}}\left(\mathbf{g}_{\bm\theta}^{-1}\left(\bfy^{(i)}\right)\right) \right|\label{eq:logProbKL}.
\end{equation}
It can be shown that this is equivalent to minimising the {KL-divergence} between the empirical distribution
$\mathcal{D}$ and the transformed distribution $\bfY=\mathbf{g}_{\bm\theta}(\bfZ)$ \citep[see, for example,][]{normalizing_flows}: %:
\begin{align}
    \argmax_{\bm\theta} \log p(\mathcal{D}| \bm\theta)
    &= \argmax_{\bm\theta}\sum_{i=1}^N \log p_\bfY \left(\bfy^{(i)}\big|\bm\theta \right) %\\
    %&=\frac{1}{N}  \sum_{i=1}^N \log p_{\mathcal{D}}\left(\bfy^{(i)}\right)
    %    +\argmax_{\bm\theta}\frac{1}{N} \sum_{i=1}^N \log p_\bfY \left(\bfy^{(i)}\big|\bm\theta \right) \\
    %&= \argmin_{\bm\theta}\frac{1}{N}  \sum_{i=1}^N \log p_{\mathcal{D}}\left(\bfy^{(i)}\right)
    %-\frac{1}{N} \sum_{i=1}^N \log p_\bfY \left(\bfy^{(i)}\big|\bm\theta \right) \\
    %&= \argmin_{\bm\theta}\sum^{N}_{i=1}  p_\mathcal{D}\left(\bfy^{(i)}\right)  \log p_{\mathcal{D}}\left(\bfy^{(i)}\right) \\
    %&\qquad- \sum_{i=1}^N  p_\mathcal{D}\left(\bfy^{(i)}\right) \log p_\bfY \left(\bfy^{(i)}\big|\bm\theta \right) \\
    = \argmin_{\bm\theta} \mathrm{KL}\left\{\mathcal D\ \vert\vert\ \mathbf{g}_{\boldsymbol{\theta}}(\mathbf Z)\right\}.
   % = \argmin_{\bm\theta} D_{\textrm{KL}}\left(\mathcal{D}\;||\; (\bfY\mid\bm\theta)\right).
\end{align}
When training a normalizing flow model, we want to find the parameter values
$\bm\theta$ that minimize the above {KL-divergence}.
This is done using back-propagation,
where the criterion $\mathcal{L}$ is set to be the negation
of Equation~\eqref{eq:logProbKL}. That is, the loss becomes the negative log likelihood of a batch
of samples from the training dataset. To perform optimisation with this criterion,
we need to compute all the terms in Equation~\eqref{eq:logProbKL}, and this expression needs to
be differentiable because the back-propagation algorithm uses the gradient of the loss
with respect to the input data.
We therefore need to find
\begin{enumerate}[(i)]
    \item an analytic and differentiable expression for the inverse transformation
$\mathbf{g}_{\bm\theta}^{-1}\left(\cdot \right)$,
    \item an analytic and differentiable expression for the {PDF} of the base distribution
$p_{\bfZ}(\cdot)$, and
    \item an analytic and differentiable expression for the log determinant of the Jacobian matrix for $\mathbf{g}_{\bm\theta}$, that is,
$\log \left|\det \mathbf{J}_{\bfZ \rightarrow {\bfY}} \right|$.
\end{enumerate}
We will derive these three components for our {EDAIN-KL} layer in the next section, %but
%before doing that, we make note of a lemma that will be used later.
%Using a result stated in \cite{normalizing_flows}, the following can be shown:
where we extensively use the following proposition \citep[see, for example,]{normalizing_flows}:
\begin{proposition}\label{thrm:normFlow}
    Let $\mathbf{g}_1,\dots, \mathbf{g}_n:\R^d \rightarrow {\R^d}$ all be bijective functions, and consider
    the composition of these functions, $\mathbf{g}=\mathbf{g}_n \circ \mathbf{g}_{n-1} \cdots \circ \mathbf{g}_1$.
    Then, $\mathbf{g}$ is a bijective function with inverse
    \begin{equation}
        \mathbf{g}^{-1}=\mathbf{g}_1^{-1} \circ \cdots \circ \mathbf{g}_{n-1}^{-1} \circ \mathbf{g}_n^{-1},
    \end{equation}
    and the log of the absolute value of the determinant of the Jacobian is given by
    \begin{equation}
        \log \left| \det \mathbf{J}_{\mathbf{g}^{-1}}(\cdot)\right|
        = \sum_{i=1}^N \log\left|\det \mathbf{J}_{\mathbf{g}_i^{-1}}(\cdot) \right|.
    \end{equation}
    %Similarly,
    %\begin{equation}
    %    \log \left| \det \mathbf{J}_{\mathbf{g}}(\cdot)\right|
    %    = \sum_{i=1}^N \log \left|\det \mathbf{J}_{\mathbf{g}_i}(\cdot) \right|.
    %\end{equation}
\end{proposition}

\subsubsection{Application to EDAIN-KL}%
\label{ssub:Application to EDAIN-KL}

Like with the {EDAIN} layer, we want to compose the outlier mitigation, shift, scale and power
transform transformations into one operation, which we do by defining
\begin{equation}\label{eq:gtheta}
    \mathbf{g}_{\bm\theta}=\mathbf{h}_1^{-1} \circ  \mathbf{h}_2^{-1} \circ \mathbf{h}_3^{-1} \circ \mathbf{h}_4^{-1},
\end{equation}
where $\bm\theta=(\bm\beta, \bm m, \bm s, \bm\lambda)$ are the unknown
parameters and
$\mathbf{h}_1,\dots,\mathbf{h}_4$ are defined in Equations~\eqref{eq:kl1}, \eqref{eq:kl2}, \eqref{eq:kl3} and \eqref{eq:kl4}, respectively.
Notice how we apply all the operations in reverse order, compared to the {EDAIN} layer. This
is because we will use $\mathbf{g}_{\bm\theta}$ to transform our base distribution $\bfZ$ into
a distribution that resembles the training dataset, $\mathcal{D}$, not the other way around.
Then, to normalize the dataset after fitting the {EDAIN-KL} layer, we apply
\begin{equation}\label{eq:gthetainv}
    \mathbf{g}_{\bm\theta}^{-1}=\mathbf{h}_4 \circ \mathbf{h}_3 \circ \mathbf{h}_2 \circ \mathbf{h}_1
\end{equation}
to each time series data point, similar to the {EDAIN} layer.
It can be shown that all the transformations defined in
Equations~\eqref{eq:kl1}, \eqref{eq:kl2}, \eqref{eq:kl3} and \eqref{eq:kl4} are invertible, of which a proof is given in
the next subsection.
Using Lemma \ref{thrm:normFlow}, it thus follows that
$\mathbf{g}_{\bm\theta}$, as defined in Equation~\eqref{eq:gtheta}, is bijective and that its inverse
is given by Equation~\eqref{eq:gthetainv}. Since we already have analytic and differentiable
expressions for $\mathbf{h}_1$, $\mathbf{h}_2$, $\mathbf{h}_3$ and $\mathbf{h}_4$ from
Equations~\eqref{eq:kl1}, \eqref{eq:kl2}, \eqref{eq:kl3} and \eqref{eq:kl4}, it follows that  $\mathbf{g}_{\bm\theta}^{-1}$, as defined in Equation~\eqref{eq:gthetainv}, is an analytic and differentiable expression, so part
(i) is satisfied.

We now move onto deciding what our base distribution should be.
As we validated experimentally in \autoref{sub:synth_dat} of the main paper,
\textit{Gaussianizing} the input data could increase the 
performance of deep neural networks (depending on the data generating process).
Therefore, we let the base distribution be the standard multivariate Gaussian distribution
\begin{equation}
    \bfZ \sim \mathcal{N}(\mathbf 0_d, \mathbf{I}_d),
\end{equation}
whose {PDF} $p_\bfZ(\cdot)$
has an analytic and differentiable expression, so part $(\textrm{ii})$ is satisfied.

%In order to optimise the unknown parameters
%$\bm\theta=(\bm\beta, \bm m, \bm s, \bm\lambda)$ of the
%{EDAIN-KL} layer, which we do by treating it as a normalizing %flow bijector, we need an analytic and
%differentiable expression for the right-hand side of %Equation~\eqref{eq:logProbKL}. We already have
%an expression for part (i) and part (ii),
In the next subsection, we will derive part (iii): an analytic and differentiable expression for
the log of the determinant of the Jacobian matrix of $\mathbf{g}_{\bm\theta}$, $\log\left|\det \mathbf{J}_{\bfZ \rightarrow {\bfY}} \right|$. Once that is done, parts (i), (ii) and (iii) are satisfied, so
$\bm\theta$ can be optimised using back-propagation
using the negation of Equation~\eqref{eq:logProbKL} as the objective. In other words, we can optimise
$\bm\theta$ to maximise the likelihood of the training data under the assumption that it comes from
the distribution $\bfY= \mathbf{g}_{\bm\theta}(\bfZ)$. This is desirable, as if we can achieve
a high data likelihood, the samples $\mathcal{D}=\{\bfy^{(i)}\}_{i=1,2,\dots,N}$
will more closely resemble a standard normal distribution
after being transformed by $\mathbf{g}_{\bm\theta}^{-1}$.
%after fitting the bijector.
%This might then increase the performance as the neural network will be fed data
%that is more Gaussian. 
Also recall that %we are working with 
multivariate time series data are considered in this work, so
the ``$\bfy$''-samples will be of the form $\bfX^{(i)} \in \R^{d\times T}$.

\subsubsection{Derivation
of the inverse log determinant of the Jacobian}%
%$\log \left|\det \mathbf{J}_{\bfZ \rightarrow {\bfX}}\right|$}%
\label{ssub:ildj}

Recall that the {EDAIN-KL} architecture is a bijector composed of four other bijective
functions. Using the result in Equation~\eqref{thrm:normFlow}, we get
\begin{equation}
    \log \left|\det \mathbf{J}_{\bfZ \rightarrow {\bfY}}(\cdot)  \right|
    = \sum^{4}_{i=1} \log \left|\det \mathbf{J}_{\mathbf{h}_i^{-1}}(\cdot) \right|.
\end{equation}
Considering the transformations in Equations~\eqref{eq:kl1}, \eqref{eq:kl2}, \eqref{eq:kl3} and \eqref{eq:kl4}, we notice that all the
transformations happen element-wise, so for $i\in\{1,2,3,4\}$, we have
$\left[\frac{\partial \mathbf{h}_i^{-1}(\bfx)}{\partial x_{k}}\right]_j =0$ for $k \neq j$.
Therefore, the Jacobians are diagonal matrices, implying that the determinant is the product of the
diagonal entries, giving
\begin{align}
    \log \left|\det \mathbf{J}_{\bfZ \rightarrow {\bfY}}(\bfx)  \right|
    %& 
    = \sum^{4}_{i=1} \log \left| \prod_{j=1}^d \left[\frac{\partial \mathbf{h}_i^{-1}(\bfx)}{\partial x_j}\right]_j   \right| %\\
    %& = \sum^{4}_{i=1} \sum_{j=1}^d \log \left| \left|\frac{\partial \mathbf{h}_i^{-1}(\bfx)}{\partial x_j}\right|_j   \right| \\
    %& 
    = \sum^{4}_{i=1} \sum_{j=1}^d \log \left| \frac{\partial h_i^{-1}\left(x_j;\theta_j^{(i)}\right)}{\partial x_j}  \right|, \label{eq:sgas9fg8a9sd8}
\end{align}
where in the last step we used the fact that $h_1,h_2,h_3$ and $h_4$ are applied element-wise
to introduce the notation $h_i(x_j;\theta^{(i)}_j)$. This means, applying $\mathbf{h}_i$ to
some vector where the $j$th element is $x_j$, and the corresponding $j$th transformation
parameter takes the value $\theta^{(i)}_j$. For example, for the scale
function $\mathbf{h}_3(\bfx)=\bfx \oslash \bm s$, we have
$h_3(x_j;\lambda_j)=\frac{x_j}{s_j}$.
From Equation~\eqref{eq:sgas9fg8a9sd8}, we know that we only need to derive the derivatives for
the element-wise inverses, which we will now do for each of the four transformations, also demonstrating that each transformation is bijective.
%In doing so, we also demonstrate that each transformation is bijective.

\paragraph{Shift.}%
\label{par:Shift}

We first consider $h_2(x_j;m_j)=x_j-m_j$. Its inverse is $h_2^{-1}(z_j;m_j)=z_j+m_j$, and it follows that
\begin{equation}
    \log \left|\frac{\partial h_2^{-1}(z_j ; m_j)}{\partial z_j} \right|
    = \log 1 = 0.
\end{equation}

\paragraph{Scale.}%
\label{par:Scale}

We now consider $h_3(x_j;s_j)=\frac{x_j}{s_j}$, whose inverse is $h_3^{-1}(x_j;s_j)={z_j}{s_j}$. It follows that
\begin{equation}
    \log \left|\frac{\partial h_3^{-1}(z_j ; s_j)}{\partial z_j} \right|
    = \log \left|{s_j}  \right|.
\end{equation}

\paragraph{Outlier mitigation.}%
\label{par:outlier mitigation}

We now consider $h_1(x_j;\beta_j)= \beta_j \tanh\left\{\frac{(x_j - \hat{\mu}_j)}{\beta_j}  \right\} + \hat{\mu}_j$. Its inverse is
\begin{equation}
    h_1^{-1}(z_j;\beta_j) =\beta \tanh^{-1} \left\{\frac{z_j - \hat{\mu}_j}{\beta_j}  \right\}
    +\hat{\mu}_j.
\end{equation}
It follows that
\begin{equation}
    \log \left|\frac{\partial h_1^{-1}(z_j ; \beta_j)}{\partial z_j} \right|
    = \log \left| \frac{1}{1-\left( \frac{z_j-\hat{\mu}_j}{\beta_j}  \right)^2}  \right|
    = -\log\left| 1-\left( \frac{z_j-\hat{\mu}_j}{\beta_j}  \right)^2 \right|.
\end{equation}

\paragraph{Power transform.}%
\label{par:Power transform}

By considering Equation~\eqref{eq:kl4}, it can be shown that for fixed
$\lambda$, negative inputs are always
mapped to negative values and vice versa, which makes the Yeo-Johnson transformation invertible.
Additionally, in $\mathbf{h}_4(\cdot)$ the Yeo-Johnson transformation is applied element-wise, so
we get
\begin{equation}
    \mathbf{h}_4^{-1}(\mathbf{z})=\left[
        \left[f_{\textrm{YJ}}^{\lambda_1}\right]^{-1}\bigg(z_1\bigg), \quad
        \left[f_{\textrm{YJ}}^{\lambda_2}\right]^{-1}\bigg(z_2\bigg), \quad \cdots, \quad
    \left[f_{\textrm{YJ}}^{\lambda_d}\right]^{-1}\bigg(z_d\bigg) \right],
\end{equation}
where it can be shown that the inverse Yeo-Johnson transformation for a single element is given by
\begin{equation}
    \left[f_{\textrm{YJ}}^\lambda\right]^{-1}\bigg(z\bigg)= \left\{
    \begin{array}{ll}
        (z \lambda + 1)^{1/\lambda} -1, & \textrm{if } \lambda \neq 0, z \geq 0; \\
        e^z-1, & \textrm{if } \lambda = 0, z \geq 0;  \\
        1-\left\{1-z(2-\lambda)\right\}^{1/ (2-\lambda)} , & \textrm{if } \lambda \neq 2, z < 0; \\
        1-e^{-z}, & \textrm{if } \lambda=2, z < 0.
    \end{array}
    \right.
\end{equation}

The derivative with respect to $z$ then becomes
\begin{equation}
    \frac{\partial \left[f_{\textrm{YJ}}^\lambda\right]^{-1}(z)}{\partial z}= \left\{
        \begin{array}{ll}
            (z \lambda + 1)^{(1-\lambda)/\lambda}, & \textrm{if } \lambda \neq 0, z \geq 0; \\
            e^z, & \textrm{if } \lambda = 0, z \geq 0;  \\
            \left\{1-z(2-\lambda)\right\}^{(\lambda-1)/(2-\lambda)} , & \textrm{if } \lambda \neq 2, z < 0; \\
            e^{-z}, & \textrm{if } \lambda=2, z < 0.
        \end{array}
    \right.
\end{equation}
It follows that
\begin{equation}
    \log \left|\frac{\partial \left[f_{\textrm{YJ}}^\lambda\right]^{-1}(z)}{\partial z} \right|= \left\{
        \begin{array}{ll}
            \frac{1-\lambda}{\lambda}\log (z \lambda + 1), & \textrm{if } \lambda \neq 0, z \geq 0; \\
            z, & \textrm{if } \lambda = 0, z \geq 0;  \\
            \frac{\lambda - 1}{2-\lambda}\log\left\{1-z(2-\lambda)\right\} , & \textrm{if } \lambda \neq 2, z < 0; \\
            -z, & \textrm{if } \lambda=2, z < 0,
        \end{array}
    \right.
\end{equation}
which we use as the expression for $\log \left| \frac{\partial h_4^{-1}\left(z_j;\lambda_j\right)}{\partial z_j} \right|$ for $z=z_1,\dots,z_d$.

%Putting all of 
Combining these expressions, we get an analytical and differentiable form
for $\log\left| \det \mathbf{J}_{\bfZ \rightarrow {\bfY}}(\bfx) \right|$, as required.

\section{EXPERIMENTAL EVALUATION DETAILS ON MODELS AND TRAINING}%
\label{sec:exp_eval_details}

In this section, we provide details on the
specific RNN model architectures used for evaluation in
\autoref{sec:experimental_evaluation} of the main paper.
We then cover the optimization procedures used for the three datasets,
including details such as number of training epochs,
learning rates and choice of optimizers.
Then, we list the learning rate modifiers used for the different
adaptive preprocessing layers, and
explain how these were selected.

\subsection{Deep Neural Network Model Architectures}

\paragraph{Synthetic dataset.}

The {GRU} {RNN} architecture consisted of two {GRU} cells with a dimensionality
of 32 and dropout layer with dropout probability $p=\frac{1}{5} $ between these cells. This was followed by
a linear feed-forward neural network with 3 fully-connected layers, separated by ReLU activation
functions, of 64, 32 and 1 units, respectively. The output was then passed through
a sigmoid layer to produce a probability $p \in (0,1)$.

\paragraph{Default prediction dataset.}

% The {GRU} {RNN} architecture is based on a ``starter model'' found on the Kaggle competition
% discussion page\footnote{
%     \url{https://www.kaggle.com/competitions/amex-default-prediction/discussion}
% } for the American Express default prediction competition, where the dataset
% originate from \citep{amex-data}. The ``starter model'' architecture was initially
% proposed by \cite{amex-starter}.

We use a {RNN} sequence model, with a classifier head, for all
our experiments with the default prediction dataset.
It consists of two stacked {GRU} {RNN} cells, both with a hidden dimensionality of 128.
Between these cells, there is a dropout layer with the dropout probability of $p=20\%$.
%Dropout is a technique used during training where each neuron is randomly deactivated with
%probability $p_{\textrm{drop}}$, which helps increase generalization performance \citep{dropout}.
For the 11 categorical features present in the dataset, we pass
these through separate embedding layers, each with a dimensionality of 4. The outputs of the
embedding layers are then combined with the output of the last
GRU cell, after it has processed the numeric columns, and the
result is passed to the linear classifier head. The classifier head is a conventional
linear neural network consisting of 2 linear layers with
128 and 64 units each, respectively, and separated by {ReLU} activation functions.
The output is then fed through
a linear layer with a single output neuron, followed by a sigmoid activation function to constrain
the output to be a probability in the range $(0,1)$. The described
architecture was chosen because it worked well in our
initial experiments.

\paragraph{Financial forecasting dataset.}

For the FI-2010 {LOB} dataset, we use
a similar {GRU} {RNN} model as with the Amex dataset, but change the architecture slightly
to match the {RNN}
model used by \cite{dain}. This was done to make the comparison between the
proposed {EDAIN} method and their {DAIN} method more fair, seeing as they also used the
{LOB} dataset to evaluate {DAIN}.
%We now describe the model architecture.
Instead of using two stacked {GRU} cells, we use one with 256 units.
We also do not need any embedding layers because all the predictor variables are numeric.
The classifier head that follows the {GRU} cells consists of one linear
layer with 512 units, followed by a {ReLU} layer and a dropout layer with a dropout probability
of $p=0.5$.  The output layer is
a linear layer with 3 units, as we are classifying the multivariate time series into one of
three classes, $\mathcal{C}=\{\textrm{decrease}, \textrm{stationary}, \textrm{increase}\}$.
These outputs are then passed to a \textit{softmax} activation function such that the
output is a probability distribution over the three classes and sums to 1: $\left[\textrm{softmax}(\bfy)\right]_j={e^{y_j}}/{\sum^{k}_{i=1} e^{y_i}},\ j=1,2,\dots,k$.
In our case, we have $k=3$.

\subsection{Optimization Procedures}\label{sub:opt_proc}

% Make sure to include the **data splits** here...

\paragraph{Synthetic dataset.}

For the synthetic dataset, all the $N_{\mathcal{D}}$ generated
datasets were separated into 80\%-20\% train-validation splits,
and the metrics reported in \autoref{tab:synth-results} of the main paper are based on
metrics computed for model performance on the validation splits.
The training was done using the Adam optimizer proposed by \cite{adam} using a base learning
rate of $\eta=10^{-3}$ and the model was trained for 30 epochs. We also used a multi-step
learning rate scheduler with decay $\gamma=\frac{1}{10}$ at the 4th and 7th epoch.
Additionally, an early stopper was used on the validation loss with a patience of 5.
 The model was trained using
binary cross-entropy loss, and the batch size was 128.

\paragraph{Default prediction dataset.}

To evaluate the different preprocessing methods on this dataset, we
perform 5-fold cross validation, which gives evaluation metrics for five different 20\% validation splits.
For training, we used a batch size of 1024.
This was chosen to give a good trade-off between the time required to train the model and
predictive performance. This model was also optimised with the Adam optimizer proposed by
\cite{adam}. %This optimizer was chosen for its reputation of being an effective optimizer in
%most cases.
The learning rate was also set to $\eta=10^{-3}$ after testing what learning rate
from the set $H=\{10^{-1}, 10^{-2}, 10^{-3}, 10^{-4}\}$ gave the most stable convergence.
The momentum parameters $\beta_1$ and $\beta_2$ were set to their default values according to
the PyTorch implementation of the optimizer \citep{pytorch}.
We also used a multi-step learning rate scheduler, with the
step size set to $\gamma=\frac{1}{10}$. For the milestones (corresponding to the
epoch indices at which the learning rate changes), we first tuned the
first step based on observing the
learning rate curve during training; then, we set the step epoch to ensure the performance did not
change too rapidly. This was done until we got milestones at 4 and 7.
We also used an early stopper based on the validation loss, with the 
patience set to $p_{\textrm{patience}}=5$ as this worked well in our
initial experiments. The maximum number of training epochs
was set to 40.

\paragraph{Financial forecasting dataset.}

As discussed in the main paper, we use an anchored cross-validation
scheme for training, which gives 9 different train-validation splits based
on the 10 days of training data available.
The targets are ternary labels $y_1,y_2,\dots,y_N \in \{0,1,2\}$, denoting whether the mid-price
decreased, remained stationary, or increased, and the output of the model is
probability vectors $\mathbf{p}_1, \mathbf{p}_2,\dots, \mathbf{p}_N \in
(0,1)^3$. For optimising the model parameters, we use the cross-entropy loss
function, defined as
\begin{equation}
    \mathcal{L}(\mathbf{p}_i, y_i)=-\sum^{2}_{c=0} \mathbb{I}_{\{y_i=c\}} \log \left( p_{i,c} \right),
\end{equation}
where $p_{i,c}$ denotes the predicted probability of class $c$ for the $i$th input sample.
We used a batch size of 128, and used the RMSProp optimizer proposed by \cite{rmsprop} to be consistent with the experiments by \cite{dain}.
The base learning rate was set to
$\eta=10^{-4}$. No learning rate scheduler nor early stoppers were used: this was done to best reproduce the methodology used by \cite{dain}.
Despite not using any early stoppers,
    all the metrics were computed based on the model state at the epoch where the validation loss
    was lowest. This was because
    the generalization performance started to decline in the middle of training in most cases.
At each training fold, the model was trained for 20
epochs. 

\subsection{Sublayer Learning Rate Modifiers}

\subsubsection{Synthetic dataset}

For the synthetic data, we use the learning rate modifier
$\eta'=10^{-1}$ for all sublayers of all adaptive preprocessing
layers (BIN, DAIN, and EDAIN), while as mentioned in Section~\ref{sub:opt_proc}, the base
learning rate is $\eta=10^{-3}$.

\subsubsection{Default prediction dataset}

The {BIN} method proposed by \cite{bin} and the {DAIN} method proposed by \cite{dain} have
both never been applied to the American Express default prediction dataset before, so their hyperparameters are tuned
for this dataset. We also tune the two versions of {EDAIN}. For all these
experiments, we run a grid-search with different combinations of the hyperparameters, specifically
the learning rate modifiers, using the validation loss from training a single model for 10
epochs on a 80\%-20\% split of the training dataset. We then pick the combination giving
the lowest validation loss. We now provide details on the grids
used for each of
the different preprocessing layers, and what learning rate
modifiers gave the best performance.

\paragraph{BIN.}%

We used the grids:

$H_{\beta}=\{10,1,10^{-1},10^{-2},10^{-6}\}$,

$H_{\gamma}=\{10,1,10^{-1},10^{-2},10^{-6}\}$, and

$H_{\lambda}=\{10,1,10^{-1},10^{-2},10^{-6}\}$.

The combination giving the lowest average cross-validation loss was found to be
$\eta_\beta=10$, $\eta_\gamma=1$, and $\eta_\lambda=10^{-6}$, giving 0.2234.

\paragraph{DAIN.}%

We used the grids:

$H_{\textrm{shift}}=\{10,1,10^{-1},10^{-2},10^{-3},10^{-4}\}$ and

$H_{\textrm{scale}}=\{10,1,10^{-1},10^{-2},10^{-3},10^{-4}\}$.

The combination giving the lowest average cross-validation loss was found to be
$\eta_{\textrm{shift}}=1$ and $\eta_{\textrm{scale}}=1$, giving $0.2216$.

\paragraph{Global-aware EDAIN.}%

We used the grids:

$H_{\textrm{scale}}=H_{\textrm{shift}}=H_{\textrm{outlier}}=H_{\textrm{pow}}=\{100,
10, 1, 10^{-1}, 10^{-2}, 10^{-3}\}$.

The combination giving the lowest average cross-validation loss was found to be
$\eta_{\textrm{shift}}=10^{-2}$,
$\eta_{\textrm{scale}}=10^{-2}$,
$\eta_{\textrm{outlier}}=10^{2}$, and
$\eta_{\textrm{pow}}=10$, giving $0.2190$.

\paragraph{Local-aware EDAIN.}%

We used the grids:

$H_{\textrm{scale}}=H_{\textrm{shift}}=\{1, 10^{-1}, 10^{-2}\}$
and
$H_{\textrm{outlier}}=H_{\textrm{pow}}=\{10, 1, 10^{-1}, 10^{-2}, 10^{-3}\}$.

The combination giving the lowest average cross-validation
loss was found to be
$\eta_{\textrm{shift}}=1$,
$\eta_{\textrm{scale}}=1$,
$\eta_{\textrm{outlier}}=10$, and
$\eta_{\textrm{pow}}=10$, giving $0.2243$.

\paragraph{EDAIN-KL.}%

Note that due to numerical gradient errors in the power transform layers occurring for some
choices of power transform learning rates, the values considered are all low to avoid these
errors. We used the grids

$H_{\textrm{outlier}}=\{100, 10, 1, 10^{-1}, 10^{-2}, 10^{-3}\}$,

$H_{\textrm{scale}}=\{100, 10, 1, 10^{-1}, 10^{-2}, 10^{-3}\}$,

$H_{\textrm{shift}}=\{100, 10, 1, 10^{-1}, 10^{-2}, 10^{-3}\}$, and

$H_{\textrm{pow}}=\{10^{-7}\}$

The combination giving the lowest average cross-validation loss was found to be
$\eta_{\textrm{shift}}=10$,
$\eta_{\textrm{scale}}=10$,
$\eta_{\textrm{outlier}}=10^{2}$, and
$\eta_{\textrm{pow}}=10^{-7}$, giving $0.2208$.

\subsubsection{Financial forecasting dataset}

The {BIN} method proposed by \cite{bin} and the {DAIN} method proposed by \cite{dain} have
already been applied to the {LOB} dataset before, but only {DAIN} has been applied
with the specific {GRU} {RNN} architecture we are using. Therefore, the learning rate
modifiers found by \cite{dain} will be used as-is ($\eta_{\textrm{shift}}=10^{-2},\eta_{\textrm{scale}}=10^{-8}$), and the learning rate modifiers for
the remaining methods will be tuned. The details of this are presented here.
For all the learning rate tuning experiments, we used the first day of data from the FI-2010
{LOB} for training and the data from the second day for validation. We then pick the
combination giving the highest validation Cohen's $\kappa$-metric.

\paragraph{BIN.}

We used the grids:

$H_{\beta}=\{10,1,10^{-1},10^{-2},10^{-6},10^{-8}\}$,

$H_{\gamma}=\{10,1,10^{-1},10^{-2},10^{-6}, 10^{-8}\}$, and

$H_{\lambda}=\{10,1,10^{-1},10^{-2},10^{-6},10^{-8}\}$.

The combination giving the highest $\kappa$ was found to be
$\eta_\beta=1$, $\eta_\gamma=10^{-8}$, and $\eta_\lambda=10^{-1}$, giving $\kappa=0.3287$.

\paragraph{Global-aware EDAIN.}%

We used the grids:

$H_{\textrm{scale}}=H_{\textrm{shift}}=H_{\textrm{outlier}}=H_{\textrm{pow}}=\{
10, 10^{-1}, 10^{-3}, 10^{-6}\}$.

The combination giving the highest $\kappa$ was found to be
$\eta_{\textrm{shift}}=10$,
$\eta_{\textrm{scale}}=10$,
$\eta_{\textrm{outlier}}=10^{-6}$, and
$\eta_{\textrm{pow}}=10^{-3}$, giving $\kappa=0.2788$.

\paragraph{Local-aware EDAIN.}%

We used the grids:

$H_{\textrm{scale}}=\{10^{-1}, 10^{-4}, 10^{-8}\}$,

$H_{\textrm{shift}}=\{10^{-1}, 10^{-2}\}$,

$H_{\textrm{outlier}}=\{10, 1, 10^{-1}, 10^{-2}, 10^{-3}, 10^{-5}, 10^{-7}\}$, and

$H_{\textrm{pow}}=\{10, 1, 10^{-1}, 10^{-2}, 10^{-3}, 10^{-5}, 10^{-7}\}$.

The combination giving the highest $\kappa$ was found to be
$\eta_{\textrm{shift}}=10^{-2}$,
$\eta_{\textrm{scale}}=10^{-4}$,
$\eta_{\textrm{outlier}}=10$, and
$\eta_{\textrm{pow}}=10$, giving $\kappa=0.3859$.

\paragraph{EDAIN-KL.}%

$H_{\textrm{scale}}=\{10^{-1}, 10^{-4}, 10^{-7}\}$,

$H_{\textrm{shift}}=\{10^{-1}, 10^{-2}\}$,

$H_{\textrm{outlier}}=\{10, 1, 10^{-1}, 10^{-2}, 10^{-3}, 10^{-5}, 10^{-7}\}$, and

$H_{\textrm{pow}}=\{10^{-2}, 10^{-3}, 10^{-5}, 10^{-7}\}$.

The combination giving the highest $\kappa$ was found to be
$\eta_{\textrm{shift}}=10^{-2}$,
$\eta_{\textrm{scale}}=10^{-4}$,
$\eta_{\textrm{outlier}}=10$, and
$\eta_{\textrm{pow}}=10^{-3}$, giving $\kappa=0.2757$.

\subsection{Other Hyperparameters}

For selecting the $\alpha$ parameter in the KDIT method,
we tried all $\alpha \in \{0.1, 1, 10, 100\}$ and selected
the value that gave the lowest average validation loss.

\subsection{Description of Computing Infrastructure}

All experiments were run on a \textit{Asus ESC8000 G4} server with the following specifications:
\begin{itemize}
    \item two 16-core CPUs of model \textit{Intel(R) Xeon(R) Gold 6242 CPU @ 2.80GHz},
    \item 896 GiB of system memory,
    \item eight GPUs of model \textit{NVIDIA GeForce RTX 3090}, each with
        $24\,576~\textrm{MiB}$ of video memory (VRAM).
\end{itemize}

\section{SYNTHETIC DATA GENERATION ALGORITHM}%
\label{sec:synth-data-appendix}

To help with designing preprocessing methods that can increase the 
predictive performance as much as possible, we propose a synthetic
data generation algorithm that gives full control over how
the variables are distributed, through only needing to specify an
unnormalized {PDF} function for each variable.   This allows synthesizing
data with distributions having the same irregularities as that seen in
real-world data.
Additionally, the covariance structure of the generated data can be
configured to resemble that of
real-world
multivariate time series, which most often have significant correlation
between the predictor variables.
Along with each time series generated, we also create a
response $y \in \{0,1\}$ that is based on the
covariates, allowing for supervised classification tasks.

%We start by providing a general overview of the algorithm, then go more in-depth into
%each part of it later.
The main input to the data generation procedure is the time series length,
$T \in \mathbb{N}$, and the number of features, $d \in \mathbb{N}$.
For each predictor variable $j=1,2,\dots,d$,
we also specify an unnormalized {PDF}, $f_j : \R \rightarrow {\R}_{+}$.
The  data generation procedure then generates a
multivariate time series covariate $\bfX \in \R^{d\times T}$ and a corresponding response $y \in \{0,1\}$ in three steps. 
Note that this procedure is repeated $N$ times to, say, generate a
dataset of $N$ time series.
An overview of the the three steps of the data generation algorithm is shown in Figure~\ref{fig:dga}. Each row in the
three matrices corresponds to one predictor variable and the column specifies the timestep.

\begin{figure}[t]
% Algorithm (equation diagram) TODO: add sample indices ^{(i)} to all the samples ?
% Also change all to lower-case to indiicate we are working with samples?
% Synth algo equation diagram {{{
\begin{align*}
    % Gaussian variable matrix left text
    \rotatebox[origin=c]{90}{{
            \parbox{3cm}{\textrm{Hidden correlated }\\
            \textrm{Gaussian RVs}}
    }}&\left\{
    \renewcommand\arraystretch{2}
    % Gaussian variable matrix
    \begin{matrix}
         N_{1,1} & N_{1,2} & \cdots & N_{1,T} \\
         N_{2,1} & N_{2,2} & \cdots & N_{2,T} \\
         \vdots & \vdots & \ddots & \vdots \\
         N_{d,1} & N_{d,2} & \cdots & N_{d,T}
    \end{matrix}\right.
    % Gaussian sample
    \;\sim\; \mathcal{N}(\mathbf{0}, \Sigma'), \textrm{ where } \Sigma' \in \R^{(dT) \times (dT)}%
    \addtocounter{equation}{1}\tag{\theequation}\label{eq:synth_algo1}
    \\%
     % Arrow down
     &\qquad\qquad\qquad\bigg\downarrow\quad U_{j,t}=\Phi_{\mathcal{N}}\left(
         N_{j,t} \big/\sqrt{\Sigma'_{jT+t,jT+t}}
     \right) \\
    % Uniform matrix text
    \rotatebox[origin=c]{90}{{
            \parbox{3cm}{\textrm{Hidden correlated }\\
            \textrm{uniform RVs}}
    }}&\left\{
    \renewcommand\arraystretch{2}
    % Uniform variable matrix
    \begin{matrix}
         U_{1,1} & U_{1,2} & \cdots & U_{1,T} \\
         U_{2,1} & U_{2,2} & \cdots & U_{2,T} \\
         \vdots & \vdots & \ddots & \vdots \\
         U_{d,1} & U_{d,2} & \cdots & U_{d,T}
    \end{matrix}\right.
    % Uniform form response
    \stackrel{\textrm{Form response}}{\xrightarrow{\hspace*{2cm} }}
    Y=\mathbb{I}\left(\sum^{d}_{j=1} \sum^{T}_{t=1} \beta_{j,t} U_{j,t}+\zeta > \frac{1}{2}  \right)
    \addtocounter{equation}{1}\tag{\theequation}\label{eq:synth_algo2} \\
     % Arrow down
     &\qquad\qquad\qquad\bigg\downarrow\quad X_{j,t}=\widehat{F^{-1}_j}\left(
         U_{j,t}
     \right) \qquad \forall t=1,2,\dots,T\\
    % Xs matrix text
    \rotatebox[origin=c]{90}{{
            \parbox{3.5cm}{\textrm{Output multivariate time series}}
    }}&\left\{
    % Xs matrix
    \renewcommand\arraystretch{2}
     \begin{matrix}
         X_{1,1} & X_{1,2} & \cdots & X_{1,T} \\
         X_{2,1} & X_{2,2} & \cdots & X_{2,T} \\
         \vdots & \vdots & \ddots & \vdots \\
         X_{d,1} & X_{d,2} & \cdots & X_{d,T}
     \end{matrix}\right.
    \addtocounter{equation}{1}\tag{\theequation}\label{eq:synth_algo3}
\end{align*}%
% }}}
\caption{High-level overview of the proposed synthetic data generation algorithm.}
\label{fig:dga}
\end{figure}

In the first step in Equation~\eqref{eq:synth_algo1}, we generate Gaussian random variables that have
a similar covariance structure to a multivariate time series. This ensures the covariates'
covariance more closely resemble that of real-world sequence data. In the second step, shown in
Equation~\eqref{eq:synth_algo2}, we convert the Gaussian random variables into uniform random variables using
the inverse normal {CDF}, after standardizing each variable.
In this step, we also form the response through a linear combination
of unknown parameters $\bm\beta$ and the uniform random variables. This ensures there is
some mutual information between the response and the covariates that are generated in the
final step.
In step 3, shown in Equation~\eqref{eq:synth_algo3}, we form the final covariates using the provided
PDFs $f_1,f_2,\dots,f_d$. This is done by estimating each {PDF}'s inverse {CDF} using
numerical methods, and transforming the uniform random variables with these. This makes the samples
come from a distribution matching that of the provided PDFs.
%We also note that the ``high-level covariance structure'' created in step 1 is maintained between
%all the steps as the transformations are all monotonic, but the magnitudes might change somewhat.
Moreover, in Equations~\eqref{eq:synth_algo1}, \eqref{eq:synth_algo2}, and \eqref{eq:synth_algo3} we use random variable notation
for each of the steps, but in practice, all the transformations are applied on samples from these.

\subsection{Step 1: Generating random variables with a time series covariance structure}%
\label{ssub:Step 1: Generating correlated random variables}

One approach to reproducing the covariance structure of a multivariate time series is to assume
that each of the $d$ individual time series follow a \textit{moving average} model, which is
a common type of theoretical time series \citep{time-series}. With this model, the covariate at
timestep $t$ takes the form
\begin{equation}\label{eq:hs08da9sd}
    X_t=c-\sum^{q}_{j=0} \theta_j \epsilon_{t-j},
\end{equation}
where $c \in \R$ is a constant, $\epsilon_0,\epsilon_1,\dots$ are uncorrelated random variables with
zero mean and finite variance $\sigma_\epsilon\in\R$. Also, $\theta_0=-1$ and
$\theta_1,\dots,\theta_q \in (-1,1)$ are the unknown parameters.
Under this model, \cite{time-series} state that the covariance between a sample
from timestep $t$ and a sample from $\tau \in \mathbb{Z}$ timesteps into the
future is 
\begin{equation}\label{eq:osmmfoieie9}
    s_\tau=\textrm{cov}\{X_t,X_{t+\tau}\}=\sigma_\epsilon^2 \sum^{q-\tau}_{j=0}
    \theta_j \theta_{j+\tau}.
\end{equation}
We will not be generating our covariates using the model in Equation~\eqref{eq:hs08da9sd} as this would
make it infeasible to get samples that are distributed according to arbitrary PDFs. However,
we can use the covariance formula from Equation~\eqref{eq:osmmfoieie9} to set the covariance between each
pair of variables generated. To do this, we first specify the parameters $q$, $\sigma_\epsilon$, and
$\theta_0,\dots,\theta_q$ for each of the $d$ predictor variables. Then, we stack the
Gaussian random variables $N_{1,1}, N_{1,2}, \dots, N_{2,1}, N_{2,2}, \dots,
N_{d,T}$ in Equation~\eqref{eq:synth_algo1} row-wise so that they form a $dT$-long vector. Let $\Sigma \in \R^{dT\times dT}$ denote the covariance matrix of this $dT$-long Gaussian multivariate random variable.
While still thinking of each $T$-length row as its own univariate time series, we fill out the 
entries in
$\Sigma$ based on Equation~\eqref{eq:osmmfoieie9}, using the parameters specified for each of the $d$
time series. The remaining entries of $\Sigma$ are randomly initialised with samples from
$\mathcal{N}(\mu=0,\sigma=\sigma_{\textrm{cor}})$, where
$\sigma_{\textrm{cor}}$ is a hyperparameter for the data synthesis, with the motivation being
to create some cross-dependence between each time series.
In order to use $\Sigma$ as a valid covariance matrix for sampling from the
$dT$-dimensional
multivariate normal distribution, it needs to be
\textit{symmetric positive semi-definite}. The $\Sigma$ matrix we have constructed so far has no
guarantee of satisfying this. Therefore, we use the algorithm proposed by
\cite{nearest_psd} to find the symmetric positive semi-definite matrix
$\Sigma' \in \R^{dT \times dT}$ that is closest to $\Sigma$ according to the Frobenius norm.
More details on this procedure can be found in \cite{nearest_psd}.
After this, we generate a $dT$-dimensional sample $\mathbf{N} \sim
\mathcal{N}(\mathbf{0}, \Sigma')$ and imagine ``unrolling'' this into a
$d\times T$ matrix where we have a $T$-timestep-long time series in each row,
just as in Equation~\eqref{eq:synth_algo1}.

\subsection{Step 2: Forming the response}%
\label{ssub:Step 2: Forming the response}

Before forming the response $y$, we need to convert the Gaussian random variables generated in
step 1 into uniform random variables. By the probability integral transform, if a normal random
variable is passed through its inverse {CDF}-function, a uniform random variable is obtained. Therefore, we do this for each of the normal random variables, as shown
in the transition between Equation~\eqref{eq:synth_algo1} and Equation~\eqref{eq:synth_algo2}, giving $d$ time series
of uniform random variables, each of length $T$.

To form the response, we randomly sample a noise term $\zeta \sim \mathcal{N}(0,\sigma_\zeta^2)$
and set
\begin{equation}
    Y=\mathbb{I}\left(\sum^{d}_{j=1} \sum^{T}_{t=1} \beta_{j,t} U_{j,t}+\zeta > \frac{1}{2}  \right).
\end{equation}
The idea behind this is to make sure each variable contributes to the response, but the contribution
of each variable might differ and some might be completely irrelevant, just like in real-world
data.
Note that the noise term $\zeta$ 
is regenerated for each multivariate time series $\bfX \in \R^{d \times T}$
we generate, while the
parameters $\bm\beta \in \R^{d \times T}$ are held fixed for each time series generated
as part of a synthetic dataset.
When synthesizing a new dataset of say $N$ samples, we first generate \textit{one set} of
$\bm\beta \in \R^{d \times T}$ unknown parameters with
$\beta_{1,1},\dots,\beta_{d,T} \stackrel{\textsc{iid}}{\sim} \mathcal{N}\left(\frac{1}{dT},\sigma_{\beta}^2\right)$, where
$\sigma_\beta$ is another hyperparameter for the dataset synthesis. The mean $\frac{1}{dT}$
was set to
ensure the dataset generated is balanced, that is, the ratio between true and false labels is equal.
% To show this, we first note that
% \begin{align}
%     \mathbb{E}\left(
%         \sum^{d}_{j=1} \sum^{T}_{t=1} \beta_{j,t} U_{j,t}+\zeta
%     \right)
%     &=\sum^{d}_{j=1} \sum^{T}_{t=1}  \mathbb{E}\left(\beta_{j,t} U_{j,t} \right)+0 \nonumber\\
%     &=\sum^{d}_{j=1} \sum^{T}_{t=1}  \mathbb{E}\left(\beta_{j,t}\right)
%     \mathbb{E}\left(U_{j,t} \right), \textrm{ as } U_{j,t}, \beta_{j,t}\textrm{ are uncorrelated}\nonumber \\
%     &=\sum^{d}_{j=1} \sum^{T}_{t=1} \frac{1}{dT} \frac{1}{2} =\frac{1}{2} \label{eq:s9sad}.
% \end{align}
% As the variables $\beta_{j,t}$s, $U_{j,t}$ and $\zeta$ are all symmetrically distributed, it can
% be shown that the term inside the $\mathbb{E}(\cdot)$ is also symmetric, so it follows from
% Equation~\eqref{eq:s9sad} that $\mathbb{E}(Y)=\mathbb{P}(Y=1)=\frac{1}{2}$. This makes the class distribution in the
% synthesized dataset balanced, which is a nice property to have.

\subsection{Step 3: Transforming the variables based on provided PDFs}%
\label{ssub:Step 3: Estimating the inverse CDF}

The third step of the data generation procedure involves taking samples from
uniformly distributed random variables $U_{j,t} \sim \textrm{U}[0,1]$ and transforming them
using the inverse {CDF} of the corresponding specified {PDF}, $f_j$, which results in
a sample from the distribution specified by $f_j$. To do this, we need to estimate the inverse
{CDF} function $\widehat{F^{-1}_j}(\cdot)$ using only $f_j$. This is done by evaluating
$f_j$ on a fine-grid of values $\mathcal{X}=\{A_j,A_j+\delta,A_j+2\delta,\dots,B_j\}$, where
$A_j$, $B_j$, and $\delta$ are additional
hyperparameters specified in conjunction with the {PDF} $f_j$.
Then we use \textit{trapezoidal numerical integration} \citep[see, for example,][]{num_anal} to estimate
$\widehat{F_j}(x)=\int_{-\infty}^x f_j(x') dx'$  for all $x \in \mathcal{X}$. Since the provided
PDFs are unnormalized, we normalize the {CDF} estimates by dividing them by
$\widehat{F_j}(B_j)$.
To get $\widehat{F_j^{-1}}(\cdot)$, we create an inverse \textit{look-up table} that maps the
$\widehat{F_j}(x)$-values to the corresponding $x$-values. When implementing this procedure,
the look-up table is \textit{cached} to ensure the integration only needs to be done
once for each $x \in \mathcal{X}$. Then, to evaluate $\widehat{F^{-1}_j}(u)$ for some $u \in (0,1)$,
we perform a \textit{binary search} \citep[see, for example,][]{algorithms} on the look-up table to find the smallest $x \in \mathcal{X}$ such that
$\widehat{F_j}(x) \geq u$, which can efficiently be done since the values are already in sorted order
because CDFs are monotonically increasing. %A reference for binary search is given by
%\cite{algorithms}.

\subsection{Hyperparameters for Synthetic Dataset Generation}

Here we present the hyperparameters used when synthesizing the irregular data we used for the experiments in
\autoref{sub:synth_dat} of the main paper. 
We used the following bounds for the three PDFs:
$(A_1,B_1)=(-8, 10)$, $(A_2,B_2)=(-30, 30)$, and $(A_3, B_3)=(-1, 7)$.
The $\theta$s were all configured with $q=3$, and we used
\begin{equation}
    \bm\Theta=\begin{bmatrix}
        \bm\theta_1 \\
        \bm\theta_2 \\
        \bm\theta_3
    \end{bmatrix}
    =
    \begin{bmatrix}
        -1 & \frac{1}{2} & -\frac{1}{5} & \frac{4}{5} \\
        -1 & \frac{3}{10} & \frac{9}{10} & 0 \\
        -1 & \frac{4}{5} & \frac{3}{10} & -\frac{9}{10} 
    \end{bmatrix}.
\end{equation}
The standard deviations were set to
$\sigma_{\textrm{cov}}=1.4$, $\sigma_{\eta}=\frac{1}{2}$, and $\sigma_\beta=2$.

\section{THE AMEX METRIC}%
\label{sec:amex_metric_sub}

Here we explain how to compute the \textit{Amex metric}
introduced in \autoref{sec:defpred}  of the main paper, which is calculated
as the mean of the default rate captured at 4\%, denoted
$D$, and the normalized Gini coefficient, referred to as $G$.
Assume we have made predictions $p_1,p_2,\dots,p_N$ for each of the $N$ customers, and assume these
predicted default probabilities have been sorted in non-increasing order. Also assume they have
associated normalized weights $w_1,w_2,\dots,w_N$ such that $\sum_{i=1}^N w_i=1$.
To compute $D$, we take the predictions $p_1,p_2,\dots,p_\omega$
captured within the highest-ranked 4\% of our predictions considering the
weights $w_1,w_2,\dots,w_N$. Then, we look at the default rate within these predictions, normalized
by the overall default rate. In other words,
\begin{equation}
    D= \frac{\sum^{\omega}_{i=1} y_i}{\sum^{N}_{i=1} y_i},
    \textrm{ where } \omega \textrm{ is the highest integer such that }
    \sum^{\omega}_{i=1} w_i \leq 0.04. %\cdot\sum^{N}_{i=1} w_i.
\end{equation}

We now describe how to compute $G$, which requires computing the Gini coefficient in two ways for
$k \in \{0,1\}$.
From \cite{gini}, we know the Gini coefficient can be computed as
\begin{equation}\label{eq:fdjf09j}
    G_k=2  \sum^{N}_{j=1} w_{i^{(k)}_j} \left(\frac{p_{i^{(k)}_j} - \overline{p}}{\overline{p}}\right)
    \left(\hat{F}_{i^{(k)}_j} - \overline{F} \right),
\end{equation}
where $\hat{F}_{i^{(k)}_j}={w_{i^{(k)}_j}}\big/{2} +\sum^{j+1}_{\ell=1}
w_{i^{(k)}_\ell}$ and $\overline{p}=\sum^{N}_{j=1} w_{i^{(0)}_j}
p_{i^{(0)}_j}=\sum^{N}_{j=1} w_{i^{(1)}_j} p_{i^{(1)}_j}$
and $\overline{F}=\sum^{N}_{j=1} w_{i^{(0)}_j}
\hat{F}_{i^{(0)}_j}=\sum^{N}_{j=1} w_{i^{(1)}_j} \hat{F}_{i^{(1)}_j} $.
To compute the normalized Gini coefficient, we first sort the predictions in non-decreasing order
by the \textit{true labels} $y_1,y_2,\dots,y_N$, and denote this ordering $i^{(0)}_1, i^{(0)}_2, \dots, i^{(0)}_N$.  Let $G_0$ denote the result of computing equation \eqref{eq:fdjf09j} with this sorting.
Then we sort the values by the \textit{predicted probabilities} $p_1,p_2,\dots,p_N$ in non-decreasing order, denoting this ordering as $i^{(1)}_1, i^{(1)}_2, \dots, i^{(1)}_N$.
Let the value of equation \eqref{eq:fdjf09j} computed with this ordering be denoted $G_1$.
The normalized Gini coefficient is then $G=G_1/G_0$, which is what we use in the final metric
\begin{equation}
    \textrm{Amex metric}=\frac{1}{2} \left(G +D \right)= \frac{1}{2} \left(\frac{G_1}{G_0}  +D \right).
\end{equation}

\end{document}